\documentclass[11pt]{article}

\usepackage[preprint]{acl}
\usepackage{amsmath}
\usepackage{placeins}
\usepackage{stfloats}
\usepackage{amssymb}
\usepackage{booktabs} % 用于三线表 \toprule, \midrule, \bottomrule
\usepackage{multirow} % 用于合并行

\usepackage{float} % 在导言区添加
\usepackage{subcaption}
\usepackage{times}
\usepackage{latexsym}

\usepackage[T1]{fontenc}
\usepackage[utf8]{inputenc}

\usepackage{microtype}

\usepackage{inconsolata}

\usepackage{graphicx}

\title{SOLAR-RL: Semi-Online Long-horizon Assignment Reinforcement Learning}

\author{
{\bfseries
Jichao Wang\textsuperscript{1}\textsuperscript{2}\textsuperscript{5}\thanks{\ \ Equal contribution} \quad
Liuyang Bian\textsuperscript{1}\footnotemark[1] \quad
Yufeng Zhou\textsuperscript{1}\textsuperscript{4}\footnotemark[1] \quad
Han Xiao\textsuperscript{1}\textsuperscript{3}
} \\
{\bfseries
Yue Pan\textsuperscript{1} \quad
Guozhi Wang\textsuperscript{1} \quad
Hao Wang\textsuperscript{1}\textsuperscript{5} \quad
Zhaoxiong Wang\textsuperscript{1}
} \\
{\bfseries
Yafei Wen\textsuperscript{1} \quad
Xiaoxin Chen\textsuperscript{1} \quad
Shuai Ren\textsuperscript{1}\thanks{\ \ Project Lead} \quad
Lingfang Zeng\textsuperscript{2}\thanks{\ \ Corresponding authors}
} \\
\textsuperscript{1}vivo AI Lab \quad
\textsuperscript{2}Zhejiang Lab \quad
\textsuperscript{3}CUHK MMLab \quad
\textsuperscript{4}Hubei University \\
\textsuperscript{5}Hangzhou Institute for Advanced Study, University of Chinese Academy of Sciences \\
\texttt{\{wangjichao23\}@mails.ucas.ac.cn}
}

\begin{document}
\maketitle
\begin{abstract}

As Multimodal Large Language Models (MLLMs) mature, GUI agents are evolving from static interactions to complex navigation.
While Reinforcement Learning (RL) has emerged as a promising paradigm for training MLLM agents on dynamic GUI tasks, its effective application faces a dilemma.
Standard Offline RL often relies on static step-level data, neglecting global trajectory semantics such as task completion and execution quality. Conversely, Online RL captures the long-term dynamics but suffers from high interaction costs and potential environmental instability. 
To bridge this gap, we propose SOLAR-RL (\textbf{S}emi-\textbf{O}nline \textbf{L}ong-horizon \textbf{A}ssignment \textbf{R}L).  Instead of relying solely on expensive online interactions, our framework integrates global trajectory insights directly into the offline learning process. 
Specifically, we reconstruct diverse rollout candidates from static data, detect the first failure point using per-step validity signals, and retroactively assign dense step-level rewards with target-aligned shaping to reflect trajectory-level execution quality—effectively simulating online feedback without interaction costs.
Extensive experiments demonstrate that SOLAR-RL significantly improves long-horizon task completion rates and robustness compared to strong baselines, offering a sample-efficient solution for autonomous GUI navigation.

\end{abstract}

\section{Introduction}
The development of autonomous agents capable of mastering Graphical User Interfaces (GUIs) is a pivotal frontier in multimodal AI. Unlike traditional agents confined to specific APIs or DOM-tree parsing, modern GUI agents powered by Multimodal Large Language Models (MLLMs) operate directly on pixel-level visual inputs, promising universal applicability across diverse operating systems \cite{qin2025ui}. While agents like CogAgent \cite{hong2024cogagent} have demonstrated proficiency in perceiving complex interfaces and executing atomic actions, a significant gap remains between "perceiving" an interface and "completing" long-horizon, multi-step workflows in real-world environments.
\begin{figure}[t]
    \centering
    % 使用 \columnwidth 让图片宽度适应单栏宽度
    \includegraphics[width=\columnwidth]{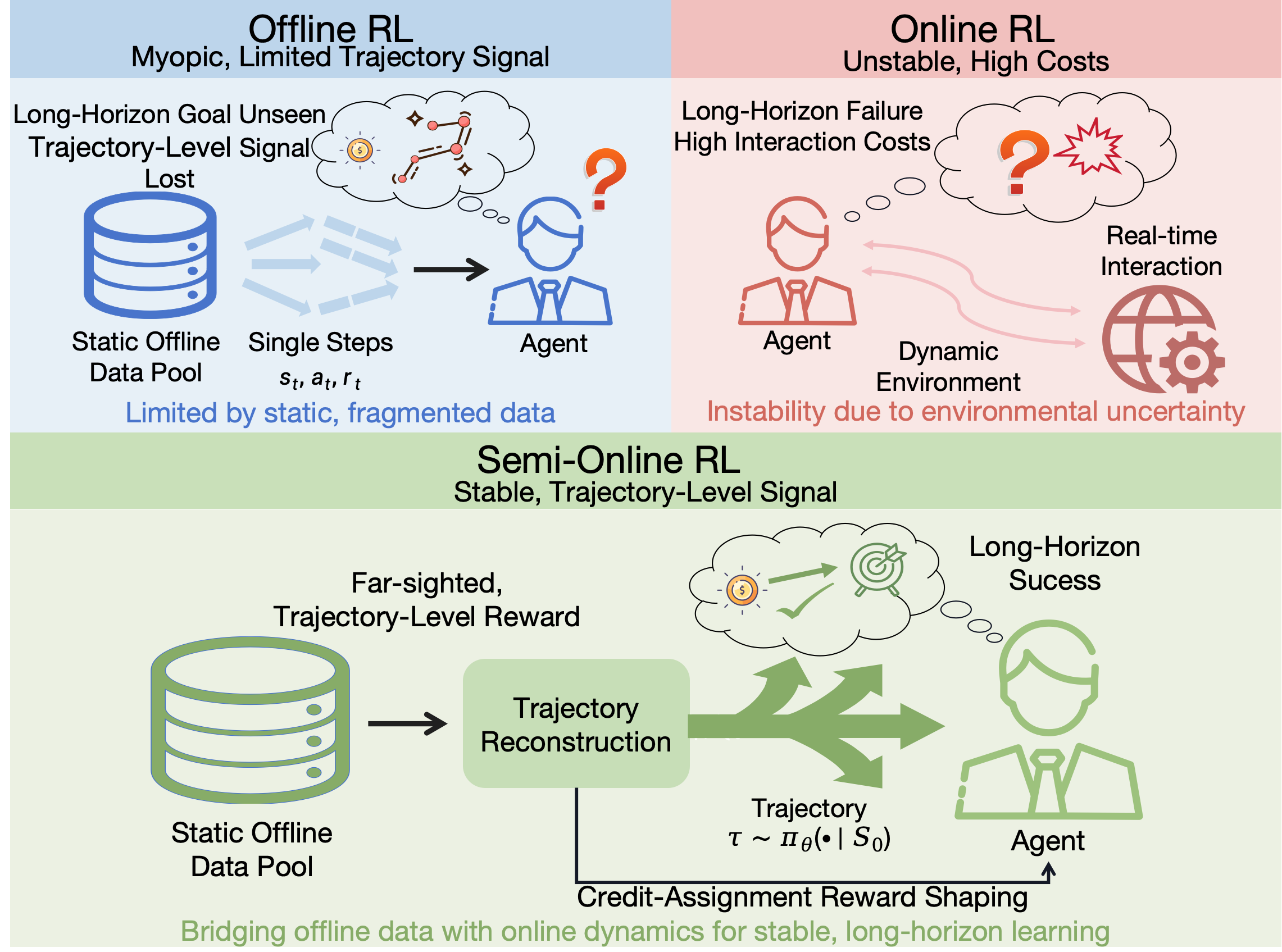}
    
    \caption{Comparison of RL paradigms for GUI agents. 
    \textbf{(Top Left)} Standard Offline RL is limited by fragmented step-level data, leading to temporal myopia and loss of global context. 
    \textbf{(Top Right)} Online RL captures dynamics but suffers from instability and prohibitive interaction costs. 
    \textbf{(Bottom)} Our SOLAR-RL bridges this gap by retrofitting global trajectory insights into offline data. It utilizes trajectory reconstruction and retroactive credit assignment via failure-point detection, combined with target-aligned reward shaping, to simulate pseudo-online feedback, ensuring stable long-horizon optimization.}
    \label{fig:framework}
\end{figure}

Current state-of-the-art (SOTA) approaches primarily rely on Supervised Fine-Tuning (SFT) or Behavior Cloning (BC) from expert demonstrations \cite{rawles2023androidinthewild, li2024effects}. While SFT effectively initializes agents with basic semantic understanding, it suffers heavily from covariate shift \cite{ross2011reduction}. When an agent encounters states deviating slightly from the training distribution—a frequent occurrence in dynamic GUIs—it lacks recovery mechanisms, leading to compounding errors. Consequently, recent works have pivoted towards Reinforcement Learning (RL) to endow agents with planning and exploration capabilities \cite{bai2024digirl, wang2025vagen}.

However, as illustrated in Figure \ref{fig:framework}, applying RL to GUI automation presents a structural dilemma, particularly for long-horizon tasks.
Online RL, while theoretically capable of capturing dynamic environmental feedback, suffers from severe instability and high variance when scaling to long trajectories. The prohibitive interaction costs and sparse reward signals in 30+ step workflows often lead to optimization failure before a successful policy can be learned~\cite{bai2024digirl}.
Conversely, Offline RL circumvents interaction risks but is plagued by \textit{temporal myopia}. By constraining learning to fragmented, step-level transitions in a static dataset, standard offline methods lose the global context required for long-term planning, making them susceptible to compounding errors~\cite{levine2020offline}.
Critically, both paradigms grapple with the Credit Assignment Problem (CAP)~\cite{lu2025ui}. In long-horizon GUI navigation, a sparse binary "success/failure" signal at the trajectory's end is insufficient to assign credit to intermediate reasoning steps. Consequently, gradients vanish or become noisy, leaving agents unable to distinguish critical decisions from irrelevant actions.

This challenge motivates the need for a paradigm that retains the stability of offline learning while incorporating trajectory-level signals typically available only in online interaction. The Semi-Online RL paradigm illustrated in Figure~\ref{fig:framework} (Right) offers a compromise. Our proposed SOLAR-RL is a concrete instantiation of this paradigm.

SOLAR-RL (\textbf{S}emi-\textbf{O}nline \textbf{L}ong-horizon \textbf{A}ssignment \textbf{R}L) synthesizes pseudo-online feedback from offline data to address the long-horizon credit-assignment problem. Specifically, we reconstruct diverse rollout candidates from static data and evaluate per-step validity to detect the first failure point where execution breaks down. We then perform retroactive credit assignment by granting positive rewards only to the valid prefix before the failure point while penalizing invalid steps after breakdown. Finally, we apply target-aligned reward shaping to align the total shaped return with trajectory-level execution quality, producing dense and stable training signals without any environment interaction.

Our main contributions are summarized as follows:
\begin{itemize} 
\item We propose SOLAR-RL, a framework that bridges the gap between offline training stability and online exploration by simulating dynamic feedback mechanisms within static datasets. 
\item We introduce a trajectory-aware reward shaping mechanism that addresses the credit assignment problem by performing retroactive credit assignment via failure-point detection and target-aligned reward shaping, distilling trajectory-level execution quality into dense step-level rewards.
\item We achieve competitive performance with online/SFT baselines without interaction, specifically demonstrating superior robustness and generalization in complex, long-horizon tasks compared to strong baselines. 
\end{itemize}

\section{Related Work}

\begin{figure*}[t!] % [t!] 表示优先尝试放在页首
    \centering
    % width=\linewidth 让图片宽度等于当前环境宽度（即两栏总宽）
    % 也可以用 width=0.9\textwidth 等比例控制
    \includegraphics[width=\linewidth]{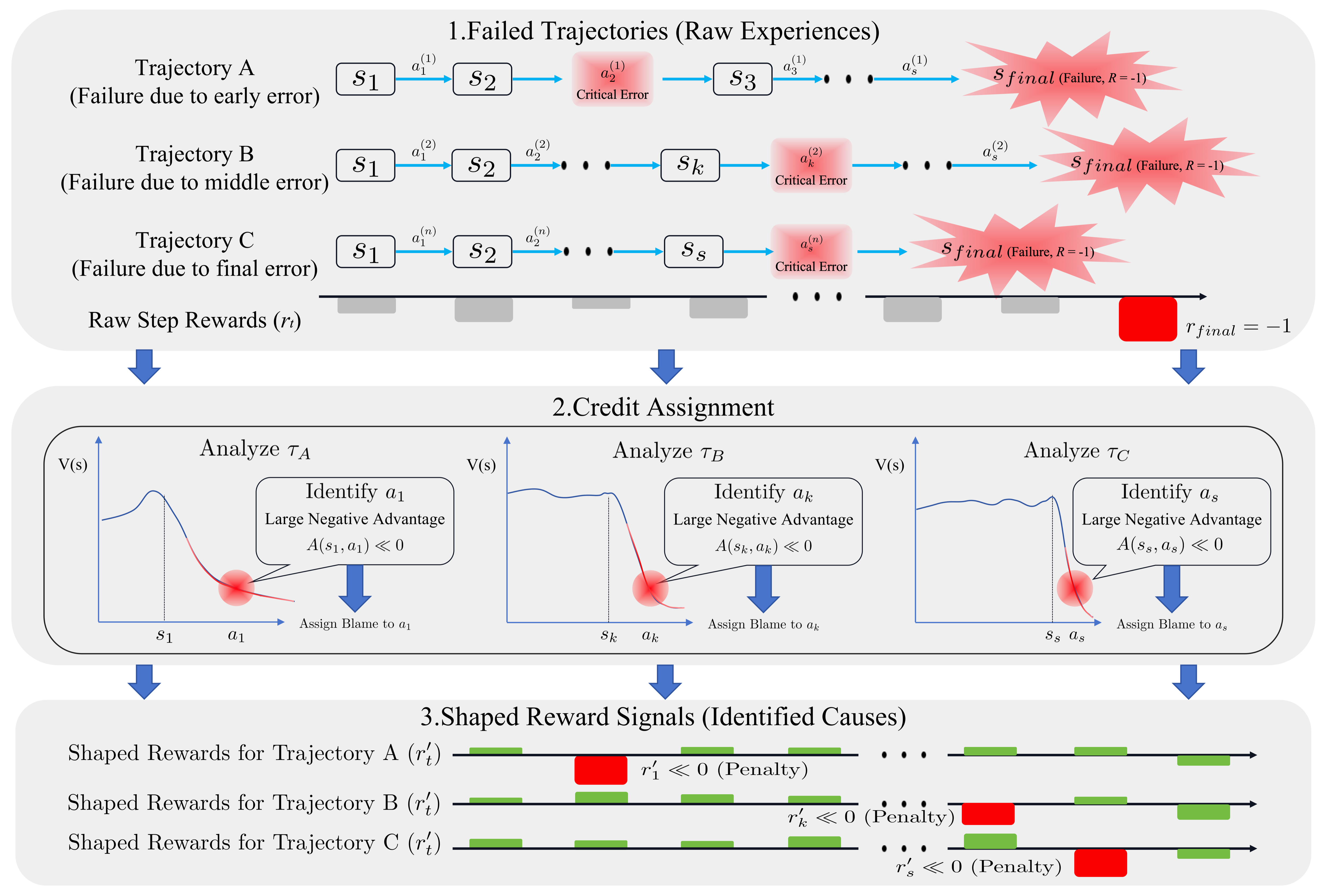}
    \caption{Illustration of the Trajectory-Aware Reward Shaping mechanism. The process consists of three stages:
    (1) \textbf{Raw Experiences:} Failed trajectories usually provide only sparse terminal feedback, obscuring where the execution first goes wrong.
    (2) \textbf{Failure-point Detection:} We identify the first breakdown step where the rollout deviates from valid execution according to per-step validity scores.
    (3) \textbf{Prefix Credit Assignment:} We assign positive rewards only to valid steps \emph{before} the breakdown, while penalizing all invalid steps along the trajectory, producing dense and stable training signals for long-horizon optimization.}
    
    \label{fig:credit_assignment_flow}
\end{figure*}

\subsection{Vision-based GUI Agents}
The evolution of GUI agents has shifted from dependency on structured metadata (e.g., DOM trees, View Hierarchies) to purely vision-based approaches that leverage MLLMs. Early works like CogAgent \cite{hong2024cogagent} introduced high-resolution visual encoders to handle small GUI elements. Recently, the field has moved towards generalist agents capable of cross-platform operation. \citet{qin2025ui} proposed UI-TARS, an end-to-end native GUI agent that achieves SOTA performance on benchmarks like OSWorld \citep{xie2024osworld} by unifying perception and action spaces.

However, while these models excel at atomic grounding, they often struggle with tasks requiring long-term planning and error correction.
A promising trend is the transition from reactive actors to deliberative reasoners. InfiGUI-R1 \cite{liu2025infigui} incorporates a "System 2" reasoning stage before action execution, significantly reducing hallucination in complex workflows. Unlike these methods that primarily focus on SFT, our work explores how to further optimize these vision-centric agents through reinforcement learning to handle long-horizon dependencies.

\subsection{Reinforcement Learning for GUI Automation} 
While SFT provides a solid initialization, it struggles to generalize to unseen states due to the discrepancy between the expert's policy and the agent's learned policy during inference. RL has been adopted to mitigate this. DigiRL \cite{bai2024digirl} pioneered an offline-to-online curriculum but faces high interaction costs in real-time environments. To address the efficiency bottleneck, Semi-Online RL has emerged as a promising paradigm. UI-S1 \cite{lu2025ui} introduced a framework that simulates online rollouts using static offline trajectories via a "Patch Module" to correct deviations, balancing stability with exploration. MobileGUI-RL \cite{shi2025mobilegui} extends this by optimizing trajectories in decentralized online environments. More recently, UI-Mem \cite{xiao2026ui} augments online GUI RL with a hierarchical experience memory that stores reusable workflows, subtask skills, and failure patterns, improving cross-task and cross-application transfer. Complementing these interaction-efficient paradigms, UI-Genie \cite{xiao2025ui} employs a specialized image-text reward model (UI-Genie-RM) to provide fine-grained, step-level supervision, enabling the iterative generation of high-quality synthetic trajectories.

Most recent approaches employ Group Relative Policy Optimization (GRPO) \cite{shao2024deepseekmath} to eliminate the need for a separate value network, reducing memory overhead. However, standard GRPO often relies on sparse trajectory-level rewards, which we argue are insufficient for long-horizon GUI tasks. Our method builds upon the semi-online paradigm but introduces a more granular, outcome-aware credit assignment mechanism.

\subsection{Credit Assignment in Long-Horizon Tasks}
The Credit Assignment Problem (CAP) remains the central challenge in RL-based GUI agents, as supervisory signals are often delayed and non-informative until the episode terminates. Recent works have attempted to densify rewards through Process Reward Models (PRMs).Beyond reward shaping, recent work also explores densifying supervision through training-only distillation. Skill-SD~\cite{wang2026skill} summarizes successful trajectories into compact natural-language skills and uses them as dynamic privileged information for teacher-guided token-level supervision, improving stability in multi-turn agent RL. Compared with such distillation-based dense supervision, SOLAR-RL focuses on trajectory-aware reward shaping and retroactive credit assignment
directly on semi-online GUI trajectories.

Most notably, VAGEN \cite{wang2025vagen} formulates visual agent tasks as Partially Observable Markov Decision Processes (POMDPs) and proposes Bi-Level General Advantage Estimation (Bi-Level GAE). VAGEN explicitly rewards "World Modeling" reasoning (state estimation and transition prediction) and propagates credit at both the turn-level and token-level. Similarly, M-GRPO \cite{hong2025multi} introduces hierarchical credit assignment for multi-agent systems.

Different from VAGEN's reliance on explicit internal world modeling or the heavy annotation cost of PRMs, SOLAR-RL proposes a trajectory-aware reward shaping mechanism. We mathematically fuse global trajectory constraints with retroactive outcome analysis, constructing dense, variation-based reward signals that offer a lightweight yet effective solution to the sparsity problem in semi-online settings.

\section{Methodology}
\label{sec:method}

In this section, we introduce \textbf{SOLAR-RL}, a Semi-online Reinforcement Learning framework designed to address the credit assignment problem in long-horizon GUI navigation. We first formulate the GUI navigation task as a POMDP. Then, we detail our two core components: (1) \textbf{Offline Trajectory Reconstruction}, which simulates online interaction dynamics using static data to expand the exploration space; and (2) \textbf{Trajectory-Aware Reward Shaping}, which retroactively propagates outcome variations to accurately attribute credit in sparse-reward environments.

\subsection{Problem Formulation}
Following prior works such as UI-S1 \cite{lu2025ui} and VAGEN \cite{wang2025vagen}, we model the GUI agent's interaction as a POMDP defined by the tuple $(\mathcal{S}, \mathcal{A}, \mathcal{P}, \mathcal{R}, \gamma)$, where $\mathcal{S}$ is the state space, $\mathcal{A}$ the action space, $\mathcal{P}$ the transition function, $\mathcal{R}$ the reward function, and $\gamma \in (0,1)$ the discount factor.
At each time step $t$, the agent observes a state $s_t$ (typically a screenshot and view hierarchy) and generates an action $a_t \sim \pi_\theta(\cdot|s_t, h_t)$ based on the interaction history $h_t$. The environment transitions to $s_{t+1}$ and provides a reward $r_t$.

Unlike standard Online RL, which suffers from high interaction costs and instability, or Offline RL, which struggles with distribution shift, our approach leverages a static dataset $\mathcal{D}_{offline}$ to approximate the optimal policy. We achieve this by simulating the dynamics of online feedback, allowing the agent to simulate pseudo-online feedback in static data by detecting failure points via per-step validity scoring and applying target-aligned reward shaping for retroactive credit assignment.

\begin{figure}[t]
    \centering
    \includegraphics[width=\columnwidth]{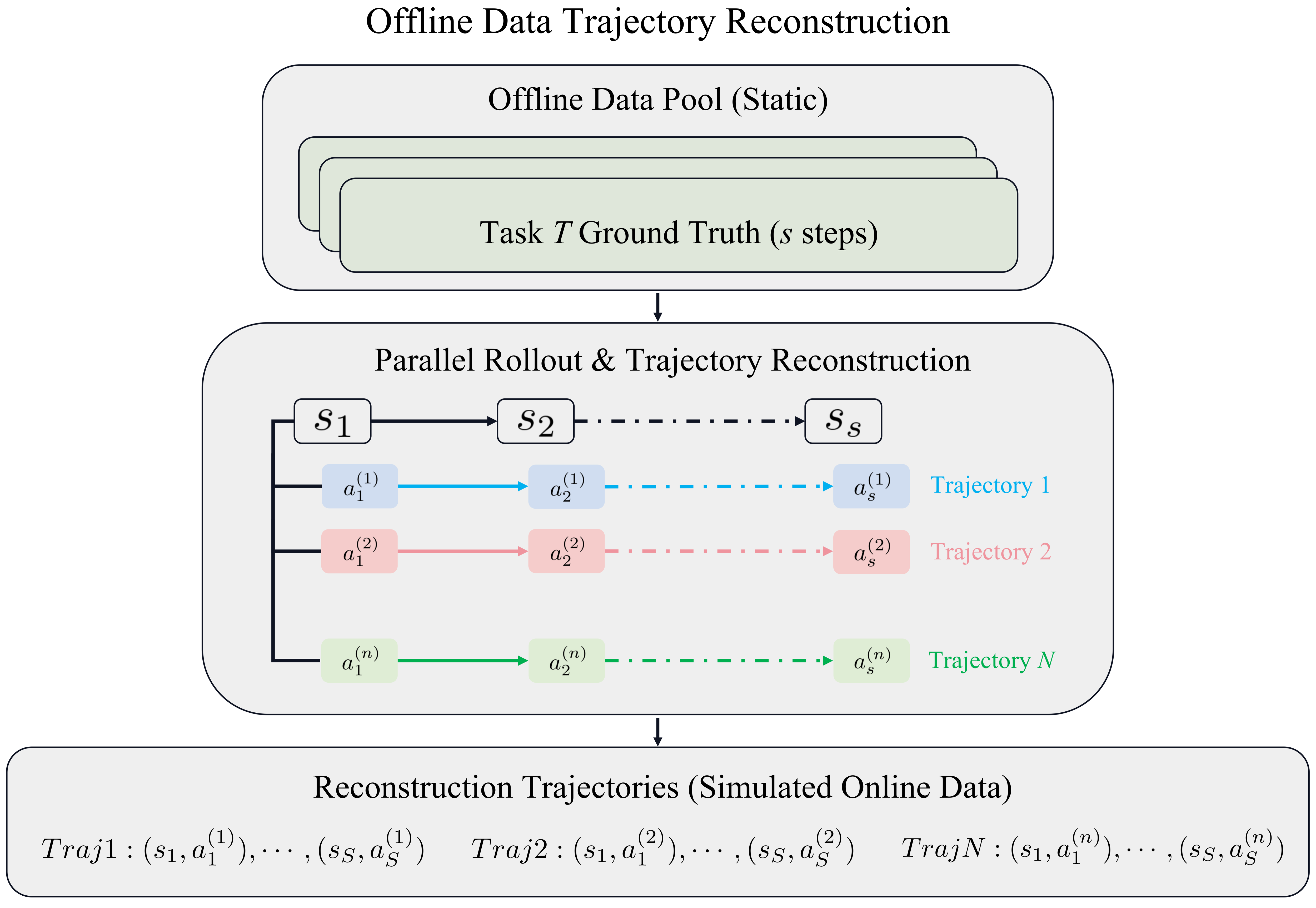}
    \caption{Offline Trajectory Reconstruction. At each step, we run $N$ parallel rollouts and connect candidates with the same rollout index to form $N$ reconstructed trajectories, yielding simulated online data for training.}
    \label{fig:traj_reconstruction}
\end{figure}

\subsection{Offline Trajectory Reconstruction}
To bridge the gap between offline training and online deployment, we propose a trajectory reconstruction mechanism, illustrated in Figure \ref{fig:traj_reconstruction}, that selectively extends valid rollout branches.
For a given task, we perform $N$ independent rollouts at each step $t$, indexed by $i \in \{1, \dots, N\}$. Responses with the same index $i$ are chained to form a potential trajectory candidate $\tau_i$. Crucially, although full rollouts are generated, a trajectory $\tau_i$ is retrospectively truncated at the first invalid step $t^*$ if the action $a_{t^*}^{(i)}$ is found to be invalid, discarding subsequent steps.

\paragraph{Validity Assessment.}
To ensure the reconstructed trajectories maintain semantic consistency with expert demonstrations, we employ a rigorous verification protocol using ground-truth (GT) labels. We categorize the supported GUI actions into four groups and apply specific metrics (e.g., Gaussian kernels for coordinates, F1-scores for text). 
The full criteria are provided in Appendix~\ref{sec:app_definitions} (Table~\ref{tab:action_criteria}).
This validity filter prunes low-quality deviations while expanding static data into diverse semi-online trajectories.

\subsection{Trajectory-Aware Reward Shaping}

A core contribution of SOLAR-RL is the Trajectory-Aware Reward Shaping mechanism (as illustrated in Figure~\ref{fig:credit_assignment_flow}), which retroactively propagates outcome variations to accurately attribute credit in sparse-reward environments.

\paragraph{Atomic Action Scoring.}
Before applying reward shaping, we quantify the execution quality of each step using a fine-grained scoring function $\Phi(a_{pred}, a_{GT}) \in [0, 1]$. We map different action primitives (e.g., Click, Scroll, Type) to continuous reward signals. 

The complete scoring functions and formulas for all atomic actions are detailed in Appendix \ref{sec:app_definitions} (Table \ref{tab:atomic_scoring}).

\subsubsection{Trajectory-Level Reward}
The trajectory-level reward $R_{traj}$ assesses the overall quality of a reconstructed trajectory. It is designed to encourage not just task completion, but also trajectory completeness and high execution quality. The formulation is defined as:
\begin{equation}
    R_{traj} = \frac{\sum_{t=0}^{T} s_{t}^{\text{raw}}}{T} + \frac{T}{N_{ref}} + \mathbb{I}(Success)
    \label{eq:trace_reward}
\end{equation}
where $s_{t}^{\text{raw}}$ is the raw validity score of step $t$, $T$ is the current trajectory length, $N_{ref}$ is the reference expert length, and $\mathbb{I}(Success)$ is the binary task completion indicator.
This reward $R_{traj}$ is assigned to every step $t$ within the trajectory. By incorporating the term $\frac{T}{N_{ref}}$, we explicitly model the trade-off between trajectory progress and task complexity, ensuring that local actions remain aligned with the global objective.

\subsubsection{Step-Level Reward with Target Alignment}

Standard step rewards often fail in long-horizon tasks due to credit assignment ambiguity and inconsistent magnitudes. To address this, we implement a Target-Aligned Reward Shaping mechanism.

\paragraph{Breakdown Step and Prefix Credit.}
Due to the trajectory truncation in Section~3.2, each reconstructed trajectory has a first breakdown step $t^*$ where the predicted action becomes invalid under our validity criteria.
We treat steps $\{0,\dots,t^*-1\}$ as the \emph{valid prefix} and only assign positive credit within this prefix.
Invalid steps (including the breakdown step) are penalized to provide explicit corrective signals.

\paragraph{Base Score Calculation.} 
First, we convert the raw validity score $s_{t}^{\text{raw}} \in [0, 1]$ into a signed base score $s_t$. Valid actions retain their positive score, while invalid actions are penalized as $-(1 - s_{raw})$. This ensures that the agent receives explicit negative feedback for deviations.

\paragraph{Target Alignment Process.}
The shaping process then harmonizes these local scores with a global budget in three phases:

1. \textbf{Aggregation:} We compute the aggregate positive score over the valid prefix and the aggregate absolute negative score:
\[
S_{\text{pos}}=\sum_{t<t^*,\, s_t>0}s_t,
\qquad
S_{\text{neg}}=\sum_{s_t<0}|s_t|.
\]

2. \textbf{Base Normalization:} We calculate a normalized reward $r_t^{base}$. 
For negative steps (errors), we impose a length-aware dynamic penalty to suppress "reward farming" in long sequences:
\begin{equation}
    r_{t}^{\text{base}} = - \left( \frac{|s_{t}|}{S_{\text{neg}}+\epsilon} + \lambda \frac{n_{\text{err}}}{\overline{T}} \right), \quad \text{if } s_t < 0
    \label{eq:step_neg}
\end{equation}
where $\lambda$ is the penalty coefficient, $n_{err}$ is the count of negative steps, and $\overline{T}$ is the batch average length. 
For positive steps in the valid prefix, we perform initial normalization:
$r_{t}^{base} = s_t / (S_{pos}+\epsilon)$ for $t<t^*$ and $s_t>0$.

3. \textbf{Total Reward Alignment:} Finally, we enforce a global constraint. We dynamically define the target total reward $R_{target}$ based on the trajectory's global quality (Eq.~\ref{eq:trace_reward}). We then compute the reward gap $\Delta = R_{target} - \sum_{t} r_{t}^{base}$ and redistribute it equally among positive steps in the valid prefix ($n_{pos}=\sum_t \mathbb{I}[t<t^* \land s_t>0]$):
\begin{equation}
    r_{t}^{final} = \begin{cases} 
    r_{t}^{base} + \frac{\Delta}{n_{pos}} & \text{if } t<t^* \land s_t > 0 \\
    r_{t}^{base} & \text{otherwise}
    \end{cases}
    \label{eq:step_final}
\end{equation}
This mechanism aligns the total return with trajectory-level quality while concentrating positive credit on \emph{pre-breakdown} decisions, thereby stabilizing long-horizon optimization under sparse feedback.

\section{Experiments}
\label{sec:experiments}

\begin{table*}[t]
\centering
\small
\setlength{\tabcolsep}{3.5pt}
\renewcommand{\arraystretch}{1.2}
\resizebox{\textwidth}{!}{%
\begin{tabular}{lcccccccccc}
\toprule
\multirow{2}{*}{\textbf{Model}} 
& \multicolumn{2}{c}{\textbf{Android Control (Low)}} 
& \multicolumn{2}{c}{\textbf{Android Control (High)}} 
& \multicolumn{2}{c}{\textbf{GUI-Odyssey}} 
& \multicolumn{1}{c}{\textbf{Android World}} 
& \multicolumn{1}{c}{\textbf{Online Data}} 
& \multicolumn{2}{c}{\textbf{Training Data}} \\
\cmidrule(lr){2-3} \cmidrule(lr){4-5} \cmidrule(lr){6-7} \cmidrule(lr){8-8} \cmidrule(lr){9-9} \cmidrule(lr){10-11}
& \textbf{TM} & \textbf{SR} 
& \textbf{TM} & \textbf{SR} 
& \textbf{TM} & \textbf{EM} 
& \textbf{SR} 
& \textbf{Online?} 
& \textbf{\#Steps} & \textbf{\#Trajs} \\
\midrule

% =================================================
% Category I: Foundation Models
% =================================================
\multicolumn{11}{l}{\textit{\textbf{I. Generalist Foundation Models} (not specifically optimized for GUI control)}} \\
Qwen2.5-VL-7B & \textbf{94.61} & \textbf{85.05} & \textbf{73.46} & \textbf{61.40} & 61.89 & 47.92 & -- & -- & -- & -- \\
GLM-4.1V-Thinking & 86.09 & 80.66 & 67.31 & 53.02 & 72.76 & 42.57 & 41.7 & -- & -- & -- \\
GLM-4.5V (106B) & 86.35 & 81.37 & 71.54 & 59.15 & \textbf{75.33} & \textbf{48.90} & \textbf{57.0} & -- & -- & -- \\
\midrule

% =================================================
% Category II: Online Models
% =================================================
\multicolumn{11}{l}{\textit{\textbf{II. Online Specialized Agents} (Uses environment interaction for data collection and iterative training)}} \\
GUI-Owl-7B & 91.05 & 86.25 & 81.60 & 72.66 & 81.58 & 65.22 & \textbf{66.4} & Y & -- & -- \\
UI-TARS-7B-SFT & 98.08 & 94.81 & 85.00 & 77.99 & 86.94 & 68.82 & 33.3 & Y & -- & 145K \\
UI-TARS-72B-SFT & \textbf{98.17} & \textbf{95.05} & \textbf{86.17} & \textbf{79.37} & \textbf{89.80} & \textbf{72.27} & 46.6 & Y & -- & 145K \\
\midrule

% =================================================
% Category III: Offline Models (Your Battleground)
% =================================================
\multicolumn{11}{l}{\textit{\textbf{III. Offline Specialized Agents} (Static data only, No interaction costs)}} \\
AgentCPM-GUI-8B & 92.80 & \textbf{88.60} & 76.40 & 67.93 & \textbf{90.82} & \textbf{74.84} & -- & N & \textgreater{}470K & \textgreater{}55K \\
UI-Venus-Navi-7B & 92.17 & 86.16 & 79.05 & 68.61 & 87.30 & 71.09 & \textbf{49.1} & N & 350K & -- \\
Aguvis-72B & -- & 84.4 & -- & 66.4 & -- & -- & 26.1 & N & -- & 35K \\

\textbf{SOLAR-RL (Ours)} 
& \textbf{93.24} & 88.57 
& \textbf{79.19} & \textbf{69.27} 
& 87.60 & 68.20 
& 33.7 
& \textbf{N} 
& \textbf{94K} & \textbf{15K} \\ 
\bottomrule
\end{tabular}
}
\vspace{-2mm}
\caption{
Unified comparison on Android Control, GUI-Odyssey, and AndroidWorld.
TM/SR denote Type Match and Step Success Rate on Android Control (Low/High); GUI-Odyssey reports Type Match (TM) and Exact Match (EM); AndroidWorld reports Success Rate (SR).
Models are grouped by training paradigm: (I) generalist foundation models, (II) agents trained with online-collected interaction trajectories, and (III) offline agents trained on static data only.
\textbf{Online?} indicates whether environment interaction is used for trajectory collection during training; \textbf{\#Steps}/\textbf{\#Trajs} report training data scale when available ("--" means not reported or not evaluated).
}
\label{tab:main_benchmarks_unified}
\end{table*}

In this section, we rigorously evaluate SOLAR-RL across three diverse benchmarks. We aim to answer two key questions: (1) Does SOLAR-RL achieve competitive performance against state-of-the-art GUI agents, particularly under strict offline constraints? (2) Does the proposed semi-online mechanism effectively mitigate the training instability and policy collapse often observed in standard RL baselines?

\subsection{Experimental Setup}

\paragraph{Benchmarks.} 
We select three representative benchmarks that cover the full spectrum of GUI agent capabilities. For SOLAR-RL, we report the mean over 4 independent runs (different random seeds). Our evaluation follows the benchmarking framework proposed in \href{https://github.com/xiaomi-research/guievalkit}{guievalkit}, and we directly adopt the reported evaluation results of other models provided by this framework:

\begin{itemize}
    \item \textbf{Android Control} \cite{li2024effects}: A widely used static benchmark for evaluating atomic action execution. It is divided into \textit{Low} (simple instruction following) and \textit{High} (requiring multi-step reasoning) splits. We report Type Match (TM) and Step Success Rate (SR).
    \item \textbf{GUI-Odyssey} \cite{lu2025guiodyssey}: A dataset designed for long-horizon cross-app navigation, featuring complex workflows that span multiple applications. This benchmark tests the agent's ability to maintain context over extended trajectories.
    \item \textbf{Android World} \cite{rawles2024androidworld}: A dynamic benchmarking environment designed to evaluate autonomous agents in real-world Android applications. Unlike static benchmarks, it employs parameterized task generation and durable, system-state-based reward signals. This benchmark serves as the most rigorous test for deployment-time reliability.
\end{itemize}

\paragraph{Baselines.} We compare SOLAR-RL with leading baselines categorized into three groups:
\textbf{(I) Generalist MLLMs} like Qwen2.5-VL \cite{bai2025qwen2} and GLM-4.5V \cite{hong2025glm};
\textbf{(II) Online Specialized Agents} like UI-TARS \cite{qin2025ui} and GUI-Owl \cite{ye2025mobile}, which benefit from environment interaction;
and \textbf{(III) Offline Specialized Agents} like AgentCPM \cite{zhang2025agentcpm}, UI-Venus \cite{gu2025ui}, and Aguvis \cite{xu2024aguvis}, which share our constraint of learning from static data.

\paragraph{Implementation Details}
We implement SOLAR-RL on top of the \href{https://github.com/volcengine/verl}{verl} framework, initialize the policy from Qwen2.5-VL-7B-Instruct and train on 15k static trajectories (Appendix~\ref{sec:app_data}). Trajectory reconstruction uses temperature $1.0$ with $N=8$ candidate rollouts per step. Training runs on 32 NVIDIA L40S GPUs with a global batch size of 128 and a maximum context length of 6,144 tokens, taking \textbf{60 hours} for \textbf{650 training steps}.Detailed hyperparameters and rollout-engine settings are provided in Appendix~\ref{sec:app_train_config}.

\subsection{Main Results}

\subsubsection{Fine-Grained Grounding and Control}
We first evaluate fundamental grounding capabilities on Android Control. As shown in Table \ref{tab:main_benchmarks_unified} (Category III), SOLAR-RL demonstrates dominant performance among offline agents.

While large-scale models benefit from massive parameter counts, SOLAR-RL achieves \textbf{93.24\%} Type Match and \textbf{88.57\%} Success Rate on the \textit{Low} split, ranking second only to AgentCPM-GUI-8B (88.60\%) by a negligible margin. Crucially, on the \textit{High} split which requires multi-step reasoning, our method achieves the \textbf{highest performance} in the offline category (69.27\% SR), outperforming both UI-Venus (68.61\%) and AgentCPM (67.93\%). This indicates that our trajectory-aware credit assignment effectively prevents reasoning degradation in complex tasks, a common issue in smaller offline models.

\subsubsection{Long-Horizon Navigation}
Table \ref{tab:main_benchmarks_unified} highlights the impact of our approach on extended interactions in GUI-Odyssey. 
Standard offline methods often suffer from compounding errors. By simulating online feedback, SOLAR-RL maintains robust performance, achieving \textbf{87.60\%} Type Match, effectively bridging the gap between step-level supervision and trajectory-level goals. While AgentCPM shows slightly higher raw metrics here, it relies on a training set that is over \textbf{3x larger} (>55k trajectories vs. our 15k), further underscoring the sample efficiency of our approach.

\subsubsection{Real-World Execution}
We further evaluate SOLAR-RL on Android World, the most dynamic and challenging benchmark. The results in Table~\ref{tab:main_benchmarks_unified} reveal a significant advantage in data efficiency.

Within the Offline Category (III), SOLAR-RL achieves a success rate of \textbf{33.7\%}, securing the second-best position. While UI-Venus achieves a higher score (49.1\%), it is trained on 350k steps of data—nearly \textbf{4 times} the volume used by SOLAR-RL (94k steps).
Moreover, when compared to Online Category (II) baselines, SOLAR-RL outperforms UI-TARS-7B-SFT (33.3\%) without requiring any expensive online interaction or the massive 145k trajectory dataset used by the latter. 
This result exposes a crucial insight: raw data scale is not the only path to performance. By refining the learning signal via trajectory-aware reward shaping, SOLAR-RL achieves competitive real-world robustness with a fraction of the data budget ($\approx$10\% of baselines), offering a far more scalable solution for autonomous GUI navigation.

\subsection{Ablation: Efficacy in Direct Training}
\label{sec:ablation_direct}

To deconstruct the contribution of our reward shaping mechanism independent of the two-stage curriculum, we conduct an ablation study involving direct RL training. 
We compare two methods: standard \textbf{GRPO} (using sparse trajectory rewards) and \textbf{SOLAR-RL} (using our proposed trajectory-aware reward shaping), both trained from scratch without the first-stage atomic adaptation.
We partition the \textit{Android Control} validation set into "Long" ($L \in [6, 13]$) and "Super Long" ($L \ge 14$) buckets based on dataset statistics (see Appendix \ref{sec:app_length_ablation}). Figure \ref{fig:super_long_ablation} visualizes the training dynamics on the most challenging "Super Long" tasks.
\begin{figure}[t]
    \centering
    \begin{subfigure}[b]{0.48\columnwidth}
        \centering
        \includegraphics[width=\linewidth]{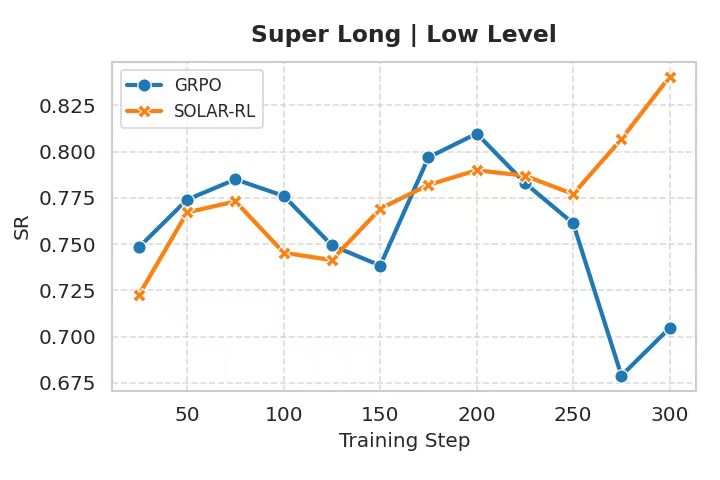}
        \caption{Low Level | Super Long}
        \label{fig:1stage_low_super_long}
    \end{subfigure}
    \hfill
    \begin{subfigure}[b]{0.48\columnwidth}
        \centering
        \includegraphics[width=\linewidth]{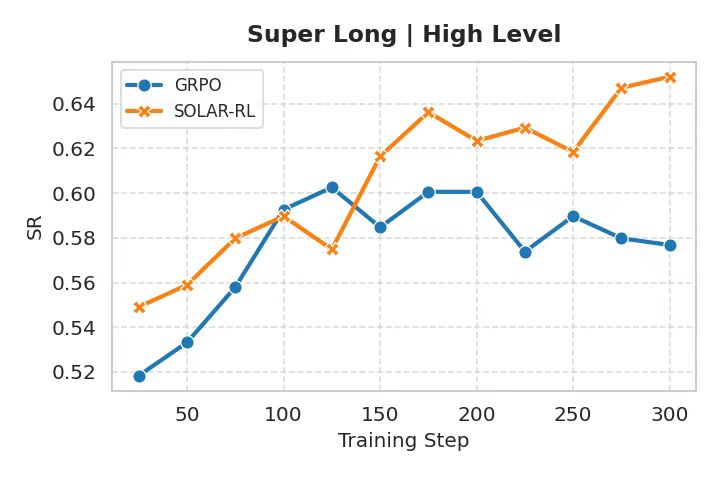}
        \caption{High Level | Super Long}
        \label{fig:1stage_high_super_long}
    \end{subfigure}
    \vspace{-2mm}
    \caption{Direct-training ablation on Super Long trajectories.
SOLAR-RL avoids late-stage collapse in GRPO and yields higher accuracy, with a larger gain on High-Level tasks.}
    \label{fig:super_long_ablation}
\end{figure}

\paragraph{Resilience in Long Horizons.}
The divergence shown in Figure \ref{fig:super_long_ablation} highlights the structural fragility of sparse-reward optimization. 
Specifically, in the Low-level split (Figure \ref{fig:1stage_low_super_long}), the baseline GRPO fails to sustain optimization after 200 steps, indicating that sparse terminal signals are insufficient to correct errors in early parts of long trajectories, leading to degenerate loops.
In contrast, SOLAR-RL leverages retroactive credit assignment to provide dense feedback, ensuring monotonic policy improvement. 
This stability advantage is further amplified in High-level tasks (Figure \ref{fig:1stage_high_super_long}), where SOLAR-RL effectively navigates complex reasoning paths that stall standard RL methods.
(Similar trends in the "Long" bucket are detailed in Appendix Figure \ref{fig:app_long_ablation}.)

\subsection{Analysis: Training Dynamics and Stability}
\label{sec:analysis}

Having verified the efficacy of our reward mechanism in direct training, we now analyze the training dynamics of the full \textbf{Two-Stage SOLAR-RL} pipeline (Atomic Adaptation $\rightarrow$ Trajectory Optimization). 
While the main results demonstrate competitive final performance, the core contribution of SOLAR-RL lies in its \textbf{training stability} and \textbf{sample efficiency}. Standard RL methods like GRPO often suffer from high variance and \textit{policy collapse} in long-horizon tasks. In this section, we provide a deep dive into the learning dynamics to validate our design choices.

\subsubsection{Training Stability}
\label{sec:analysis_actions}

\begin{figure}[t]
    \centering
    \includegraphics[width=\columnwidth]{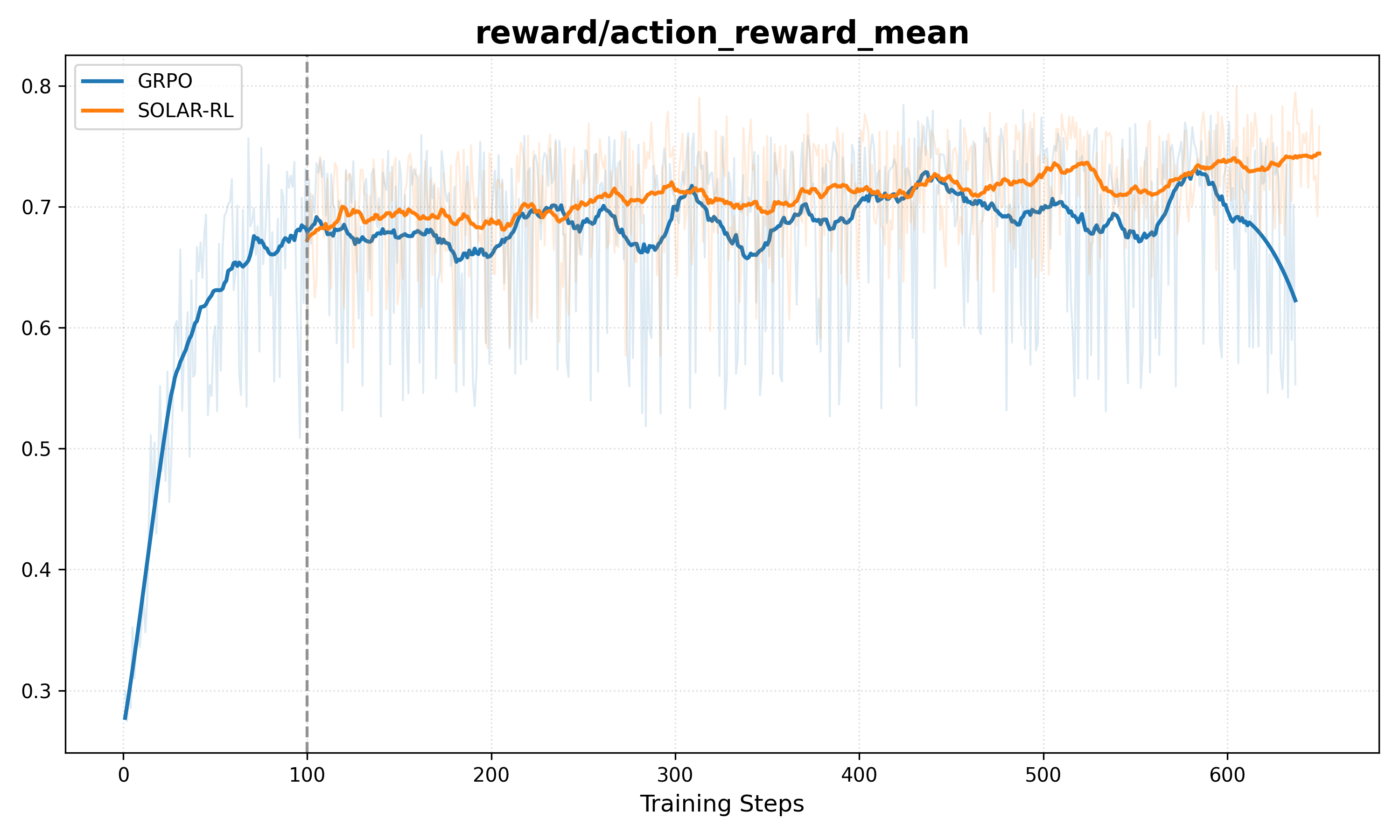}
    \caption{Mean action reward during training. GRPO collapses in later stages, whereas SOLAR-RL improves monotonically and converges.}
    \label{fig:mean_reward}
\end{figure}

We compare SOLAR-RL against the strong baseline, GRPO, to understand why our two-stage pipeline is more stable. 
Figure~\ref{fig:mean_reward} shows the evolution of the mean action reward. GRPO improves early but suffers catastrophic degradation after $\sim$600 steps, a typical \textit{policy collapse} failure mode in long-horizon RL where sparse rewards provide inconsistent gradients and encourage overfitting to local optima (e.g., loops). 
In contrast, SOLAR-RL exhibits monotonic improvement and plateaus at a higher reward ($\approx 0.75$), suggesting that the two-stage curriculum with trajectory-aware credit assignment and reward shaping stabilizes long-term optimization.

\begin{figure}[t]
    \centering
    \includegraphics[width=\columnwidth]{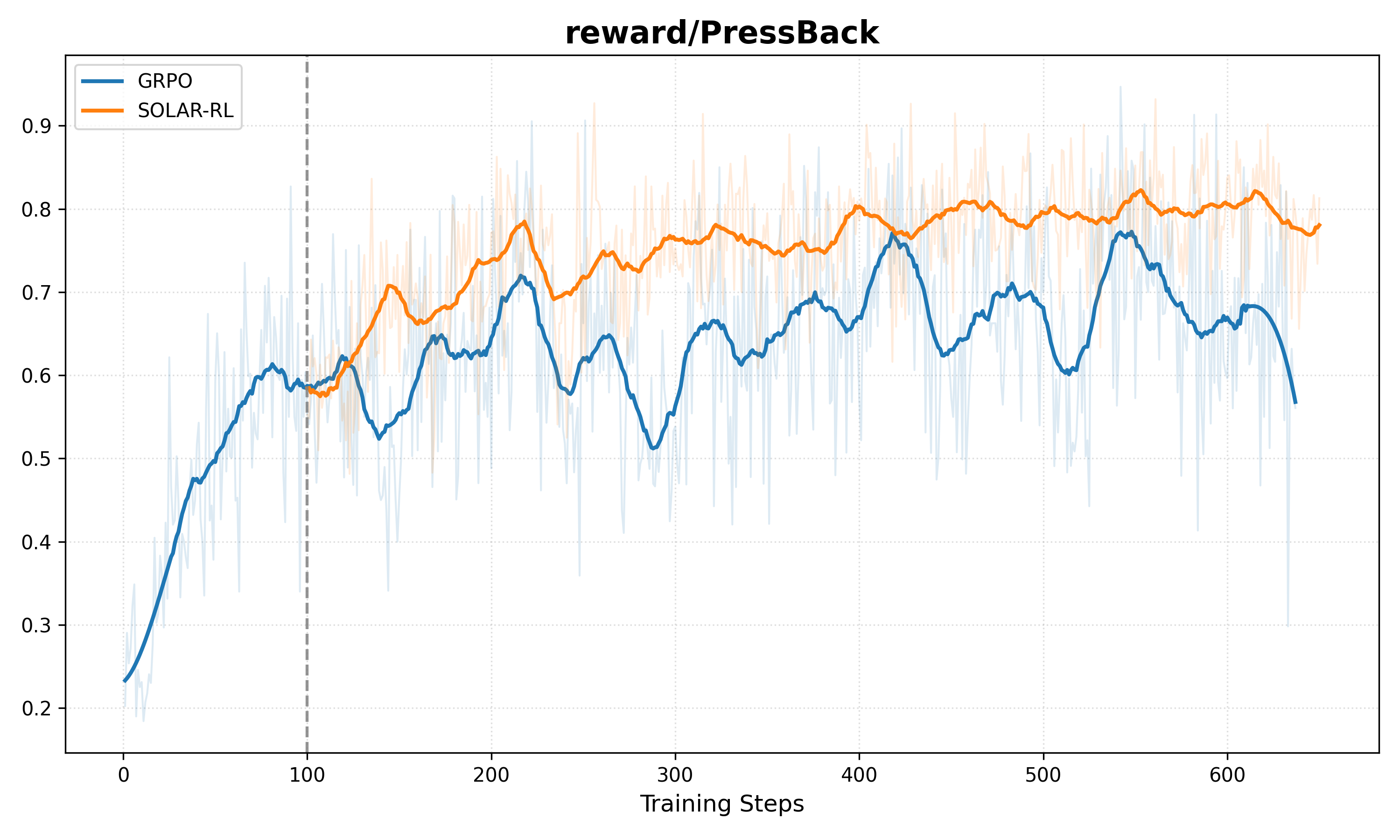}
    \caption{Training dynamics of \textit{PressBack}. SOLAR-RL converges faster and more stably than GRPO, effectively preventing navigation loops.}
    \label{fig:pressback_main}
\end{figure}

To localize the source of this stability, we further inspect action-level learning. 
Figure~\ref{fig:pressback_main} highlights \textit{PressBack}, a critical primitive for error correction.The GRPO baseline exhibits severe oscillation, indicating difficulty in learning when to retreat. 
SOLAR-RL, leveraging retroactive feedback, quickly converges to high precision ($>0.8$), confirming that our method injects global trajectory context into atomic action learning and mitigates the "forgetting" of skills during trajectory-level optimization.

\paragraph{Generalization to Other Primitives.}
Similar stabilization trends hold across other primitives; detailed per-primitive curves and discussions are provided in \textbf{Appendix~\ref{sec:app_action_ablation}} and \textbf{Figure~\ref{fig:appendix_actions}}.We further provide a qualitative failure-case study on continuous decision recovery in Appendix~\ref{sec:app_case_study}.

\subsubsection{Long-Horizon Resilience}
\label{sec:analysis_2stage_long}

Finally, we investigate whether the benefits of SOLAR-RL persist in the most challenging long-horizon scenarios even after Atomic Adaptation. We focus on the \textbf{High-Level} split across the \textit{Long} and \textit{Super Long} horizons.

\begin{figure}[t]
    \centering
    \begin{subfigure}[b]{0.48\columnwidth}
        \centering
        \includegraphics[width=\linewidth]{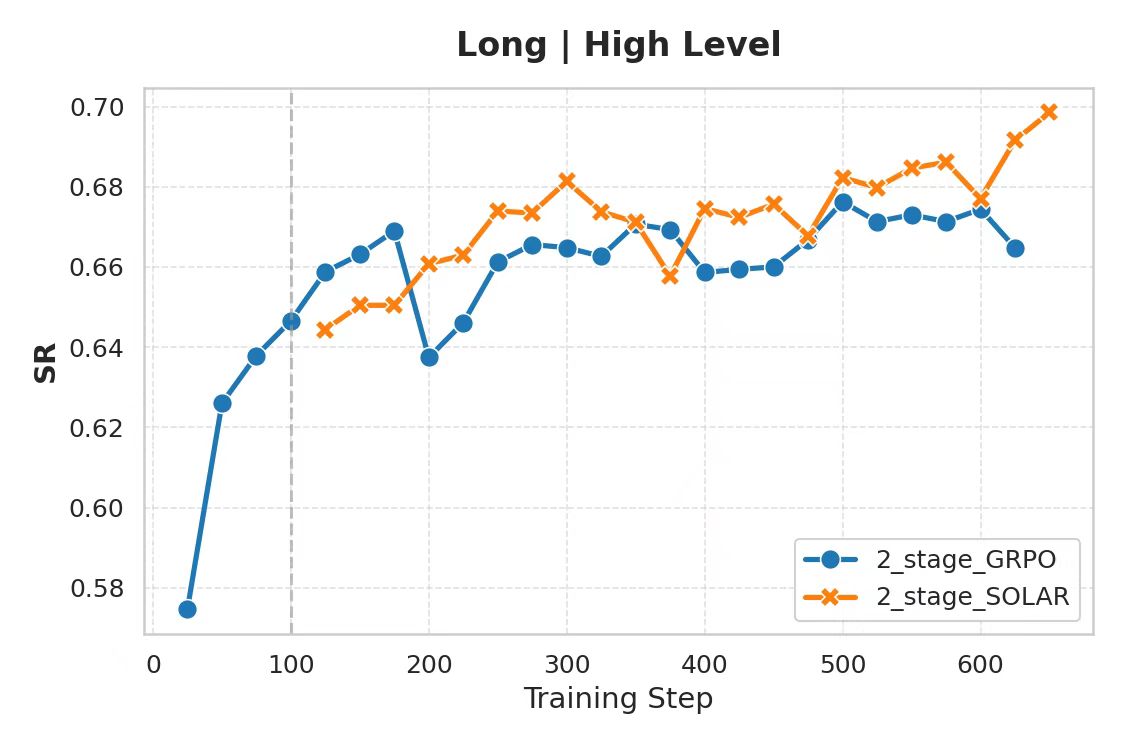}
        \caption{High Level | Long}
        \label{fig:2stage_long}
    \end{subfigure}
    \hfill
    \begin{subfigure}[b]{0.48\columnwidth}
        \centering
        \includegraphics[width=\linewidth]{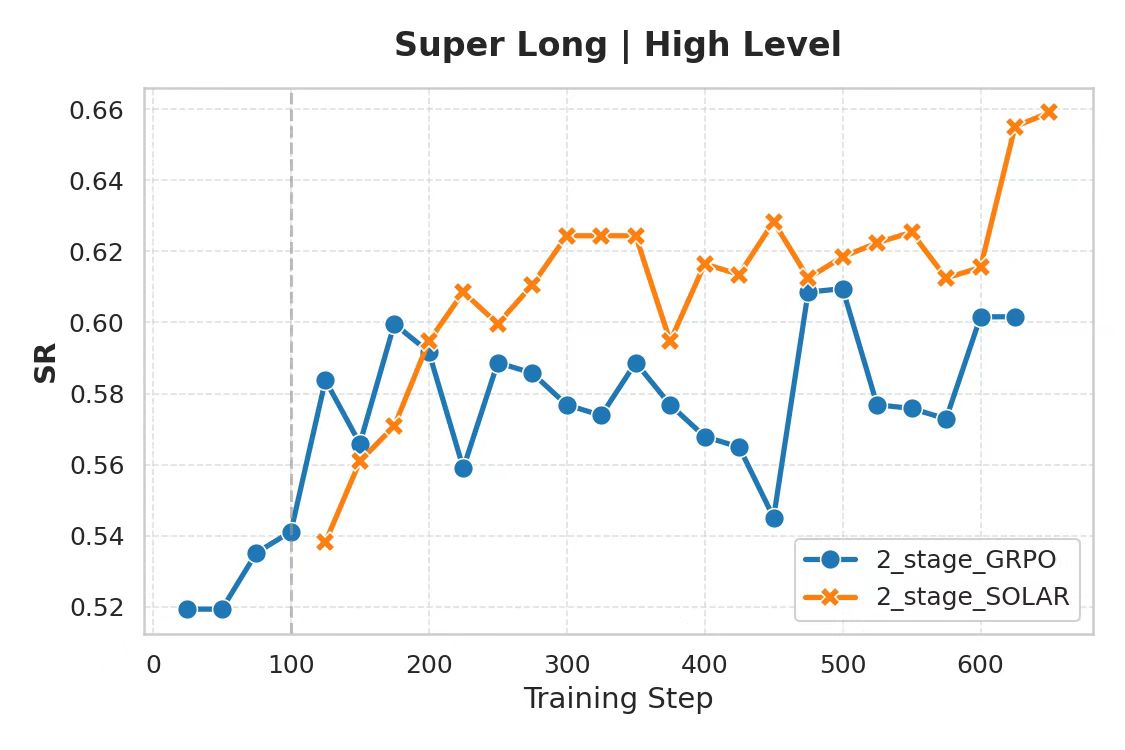}
        \caption{High Level | Super Long}
        \label{fig:2stage_superlong}
    \end{subfigure}
    \vspace{-2mm}
    \caption{Two-Stage Training Dynamics on High-Level Tasks. SOLAR-RL improves more steadily and reaches higher accuracy than GRPO.}
    \label{fig:2stage_ablation}
\end{figure}

As shown in Fig.~\ref{fig:2stage_ablation}, SOLAR-RL remains robust under the most challenging long-horizon settings.
On the \textbf{Long} horizon (Fig.~\ref{fig:2stage_long}), the 2-stage GRPO baseline quickly saturates around $0.66$--$0.67$, whereas 2-stage SOLAR-RL continues to improve and reaches $\approx 0.70$.
The advantage becomes more pronounced on the \textbf{Super Long} horizon (Fig.~\ref{fig:2stage_superlong}): the 2-stage GRPO baseline oscillates and largely plateaus around $\approx 0.58$--$0.60$, suggesting that a good initialization alone is insufficient for stable long-horizon optimization.
In contrast, 2-stage SOLAR-RL continues to improve and reaches a peak SR of $\approx 0.66$, confirming that trajectory-aware reward shaping provides critical dense guidance for long-term planning beyond atomic adaptation.

\section{Conclusion}
\label{sec:conclusion}

In this paper, we presented \textbf{SOLAR-RL}, designed to bridge the gap between static offline datasets and dynamic online decision-making for GUI agents. 
Addressing the limitations of myopic offline cloning and sample-inefficient online interaction, we introduced two core mechanisms: Offline Trajectory Reconstruction to simulate diverse interaction pathways, and Trajectory-Aware Reward Shaping to solve the credit assignment problem in sparse-reward environments.
Extensive evaluations on Android Control, GUI-Odyssey, and Android World demonstrate that SOLAR-RL improves robustness in \emph{long-horizon} tasks, preventing the policy collapse often observed in baselines.

\section{Limitations}
\label{sec:limitations}

While SOLAR-RL effectively bridges the gap between offline stability and online exploration for GUI agents, several limitations remain.

First, the \textbf{Semi-Online mechanism is bounded by the coverage of the offline dataset}. Although our trajectory reconstruction expands the exploration space by synthesizing diverse interaction paths, the agent still cannot encounter states or system dynamics that are entirely absent from the source data distribution, such as unseen pop-ups, latency-induced interface shifts, or rare app-specific workflows. In this sense, SOLAR-RL simulates online feedback, but it does not replace real interaction with a live environment. 

Second, the \textbf{current instantiation relies on ground-truth signals for trajectory validity checking}. In our experiments, the validity filter in Section~\ref{sec:method} uses expert annotations and heuristic comparisons (e.g., coordinate distance, text matching, and action-type consistency) to identify the first breakdown step in a clean and controlled manner. However, SOLAR-RL is agnostic to how the validity score is obtained: it can come from (i) ground-truth labels, as in this work, (ii) a learned verifier trained on labeled offline data to predict action validity from observation-action pairs, or (iii) a reward or critic model, such as process reward models or UI verifiers, when explicit labels are unavailable. This suggests a feasible path toward scaling SOLAR-RL to weakly labeled or unlabeled GUI data. At the same time, replacing ground-truth supervision with learned validity estimators introduces new challenges, including reward noise, calibration drift, and reward hacking, which require further investigation.

Third, our experimental evaluation is primarily \textbf{concentrated on mobile (Android) environments}. Although trajectory-aware reward shaping is platform-agnostic in principle, extending it to desktop operating systems and web browsers is non-trivial. These environments introduce richer action primitives (e.g., hover, right-click, keyboard shortcuts, drag-and-drop, multi-window interactions, and tab switching), more asynchronous interface changes, and more complex view hierarchies than mobile benchmarks. A faithful extension therefore requires not only broader benchmarks, but also platform-specific validity criteria, curated offline trajectories, and standardized training/evaluation protocols. We leave this cross-platform extension to future work.

\bibliography{custom}
\clearpage
\appendix

\section{Data Statistics}
\label{sec:app_data}

Table \ref{tab:data_statistics} summarizes the statistics of the open-source data sources and the specific subset selected for training our policy.

\begin{table}[h]
    \centering
    \small
    \resizebox{\linewidth}{!}{
    \begin{tabular}{lcccc}
        \toprule
        \textbf{Dataset} & \textbf{Android Control} & \textbf{GUI-Odyssey} & \textbf{GUI-Act} & \textbf{Selected for Training} \\
        \midrule
        \textbf{\# Steps} & 84k & 114k & 15k & 94k \\
        \textbf{\# Trajectories} & 13k & 7k & 4k & 15k \\
        \bottomrule
    \end{tabular}
    }
    \caption{Dataset statistics. "Selected for Training" denotes the 15k filtered high-quality trajectories used for initial policy optimization.}
    \label{tab:data_statistics}
\end{table}

\begin{figure*}[b]
    \centering
    \vspace{-10pt} % 稍微减少与上方文字的距离
    
    % 第一行：Launch, Finished, Scroll
    \begin{subfigure}[b]{0.32\linewidth} % 宽度设为 0.32，三张图并排刚好占满
        \centering
        \includegraphics[width=\linewidth]{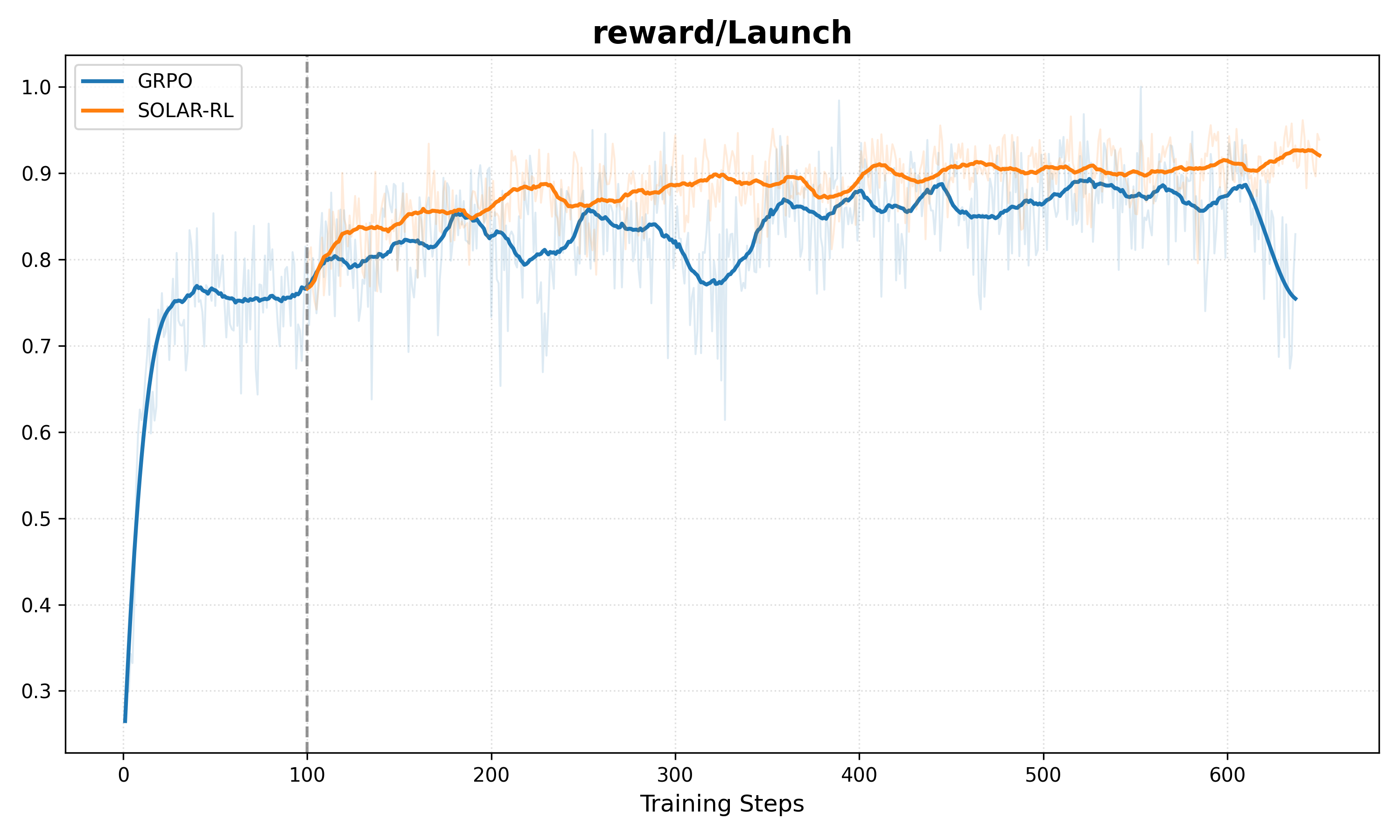}
        \caption{Launch}
        \label{fig:launch}
    \end{subfigure}
    \hfill
    \begin{subfigure}[b]{0.32\linewidth}
        \centering
        \includegraphics[width=\linewidth]{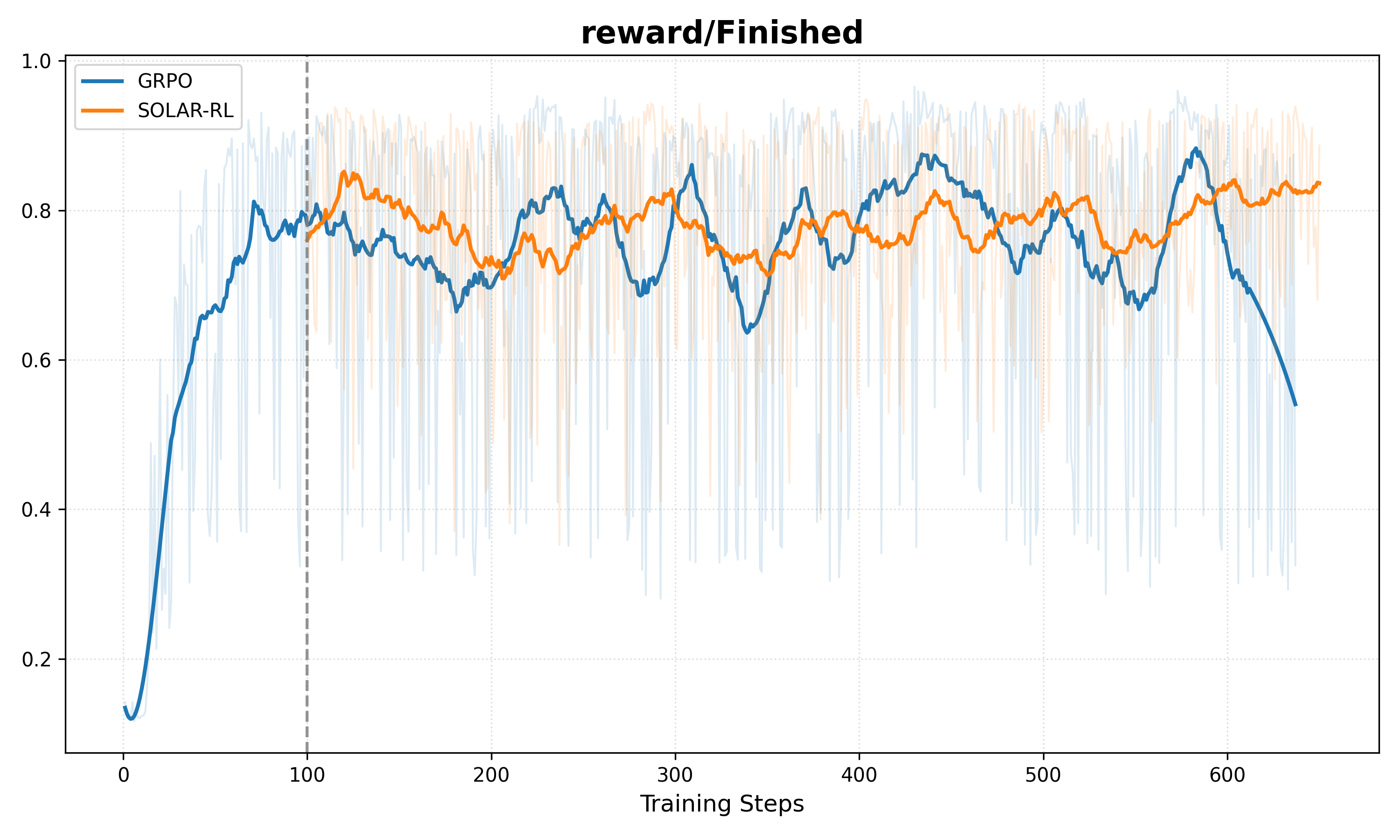}
        \caption{Finished}
        \label{fig:finished}
    \end{subfigure}
    \hfill
    \begin{subfigure}[b]{0.32\linewidth}
        \centering
        \includegraphics[width=\linewidth]{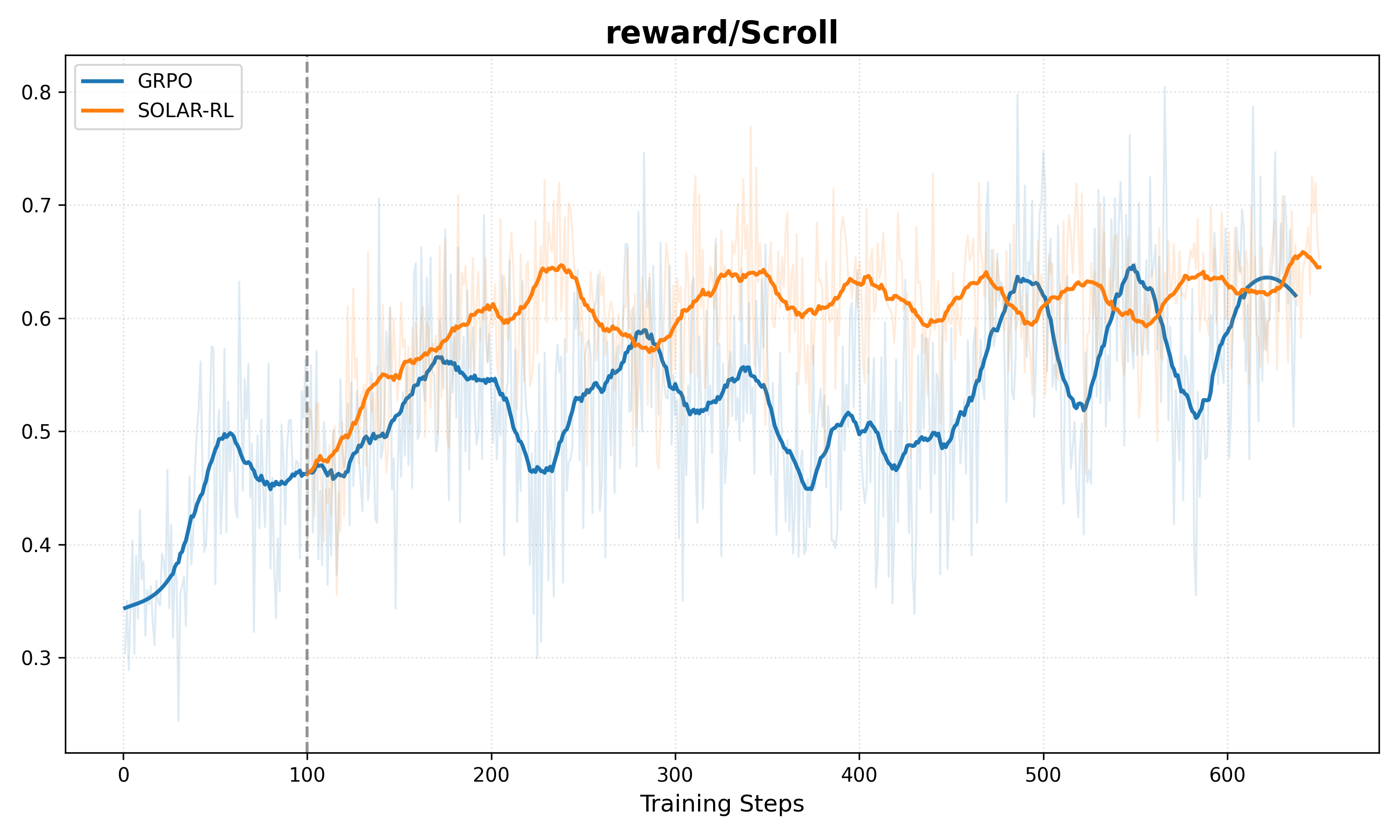}
        \caption{Scroll}
        \label{fig:scroll}
    \end{subfigure}
    
    \vspace{0.2cm} % 行间距
    
    % 第二行：Wait, Type, Click
    \begin{subfigure}[b]{0.32\linewidth}
        \centering
        \includegraphics[width=\linewidth]{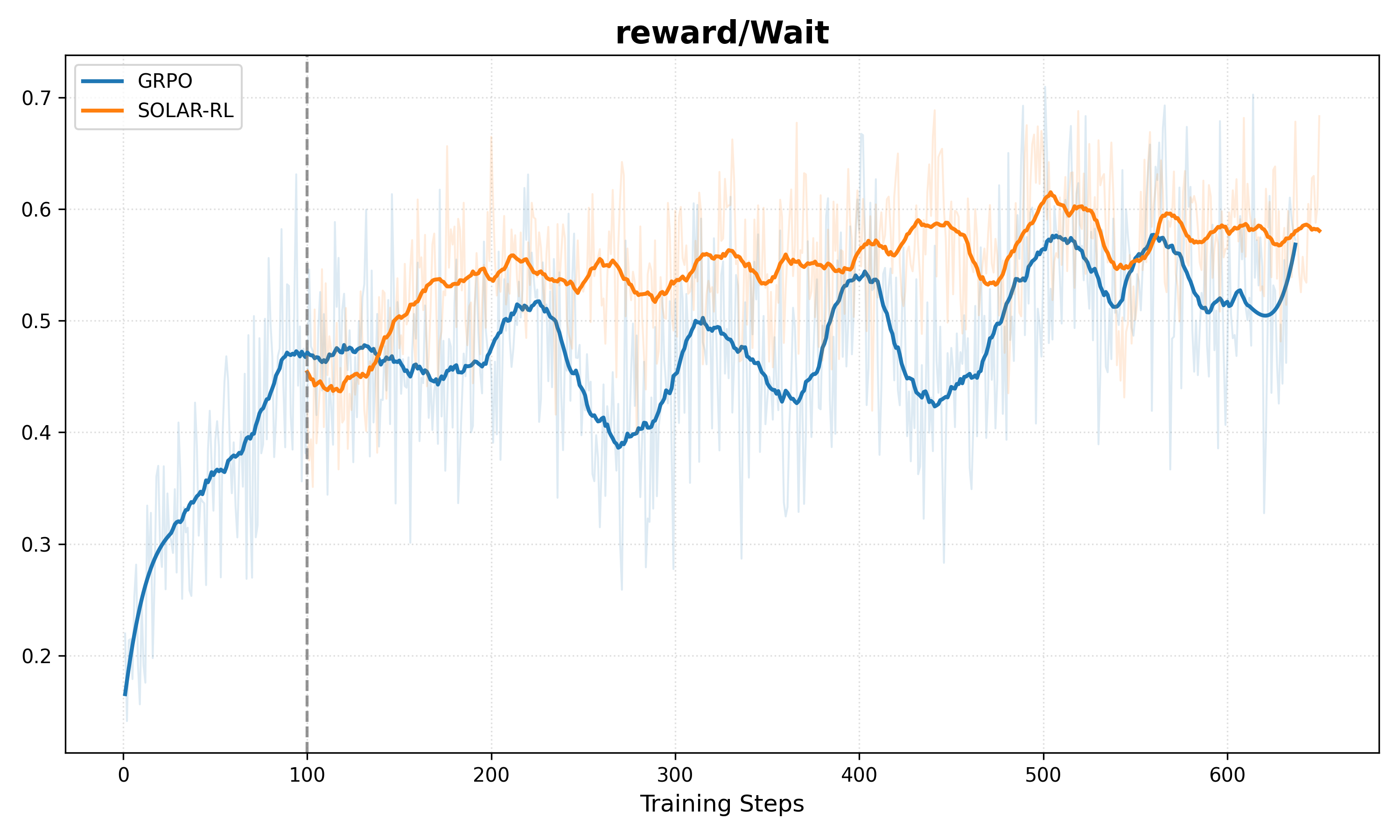}
        \caption{Wait}
        \label{fig:wait}
    \end{subfigure}
    \hfill
    \begin{subfigure}[b]{0.32\linewidth}
        \centering
        \includegraphics[width=\linewidth]{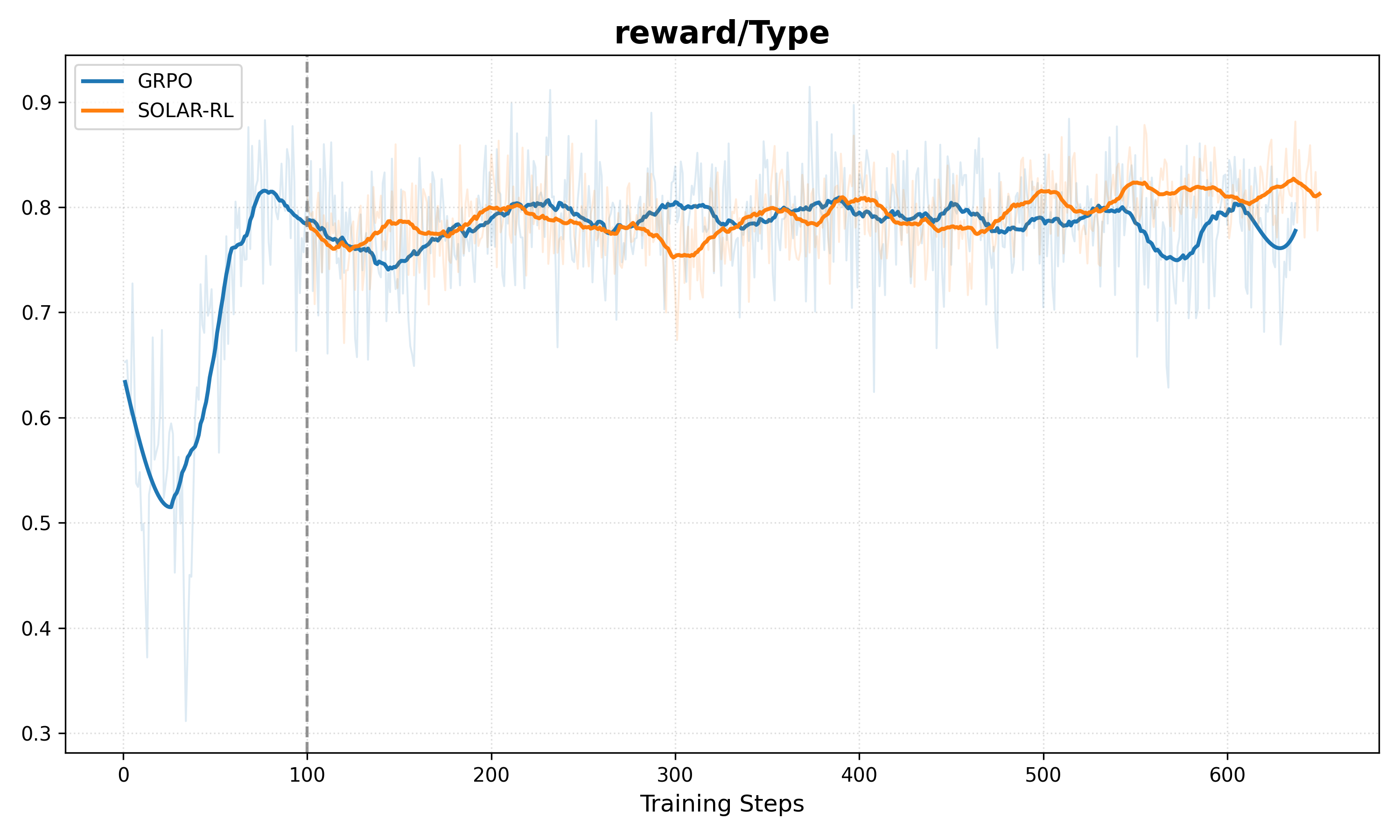}
        \caption{Type}
        \label{fig:type}
    \end{subfigure}
    \hfill
    \begin{subfigure}[b]{0.32\linewidth}
        \centering
        \includegraphics[width=\linewidth]{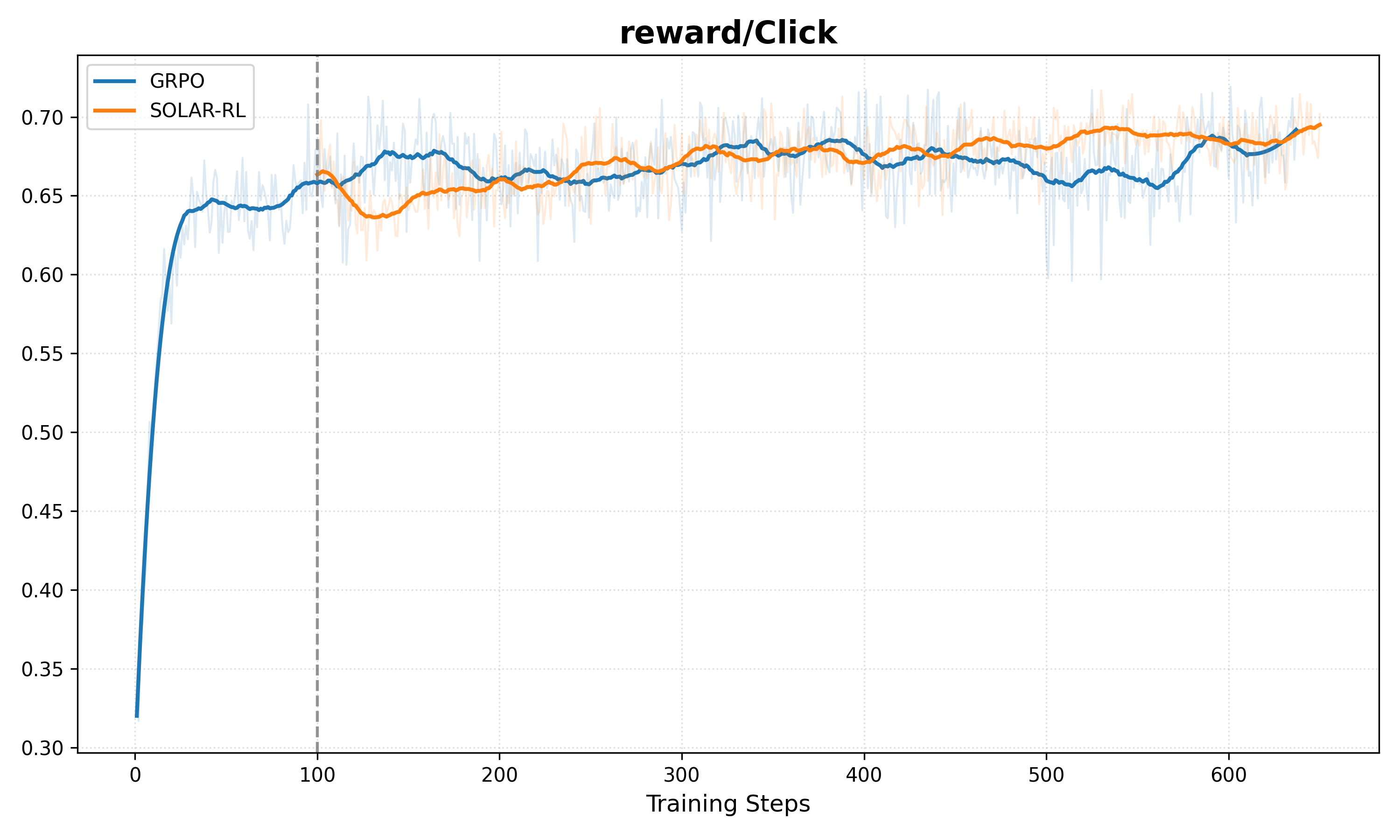}
        \caption{Click}
        \label{fig:click}
    \end{subfigure}
    
    \vspace{-0.2cm}
    \caption{Supplementary training curves for the six remaining action primitives. SOLAR-RL (Orange) demonstrates consistent stability advantages over GRPO (Blue).}
    \label{fig:appendix_actions}
\end{figure*}

\section{Implementation Details and Supplementary Materials}
\label{sec:appendix}

\subsection{Detailed Training Configuration}
\label{sec:app_train_config}

Table~\ref{tab:train_config} reports the detailed training configuration used in our experiments. Unless otherwise noted, GRPO and SOLAR-RL share the same initialization, rollout-engine settings, optimization hyperparameters, and training budget; the primary difference lies in the reward definition.

\begin{table*}[t]
  \centering
  \small
  \renewcommand{\arraystretch}{1.08}
  \setlength{\tabcolsep}{8pt}
  \begin{tabular}{p{0.18\textwidth}p{0.34\textwidth}p{0.36\textwidth}}
      \toprule
      \textbf{Category} & \textbf{Hyperparameter} & \textbf{Value} \\
      \midrule
      Initialization & Base model & Qwen2.5-VL-7B-Instruct \\
      Rollout & Temperature / $N$ & 1.0 / 8 candidates per step \\
      Context & Max context length / max response length & 6144 / 1024 \\
      Optimizer & Actor learning rate / $\gamma$ & $1\times10^{-6}$ / 0.95 \\
      Optimizer & PPO clip ($\texttt{clip\_ratio\_low/high}$) & 0.2 \\
      Regularization & KL coefficient / KL loss type & 0.001 / \texttt{low\_var\_kl} \\
      Budget & Train batch size / update steps & Global batch size = 128; 650 update steps ($\sim$60h) \\
      Compute & GPUs & 32$\times$ NVIDIA L40S \\
      Rollout/Engine & \texttt{ppo\_max\_token\_len\_per\_gpu} & 16384 \\
      Rollout/Engine & \texttt{log\_prob\_max\_token\_len\_per\_gpu} & 25600 \\
      Rollout/Engine & \texttt{max\_num\_batched\_tokens} & 32768 \\
      Rollout/Engine & \texttt{limit\_images} & 1 \\
      Rollout/Engine & \texttt{rollout.name} & \texttt{vllm} \\
      Rollout/Engine & \texttt{VLLM\_ATTENTION\_BACKEND} & \texttt{XFORMERS} \\
      \bottomrule
  \end{tabular}
  \caption{Detailed training configuration used for SOLAR-RL and the matched GRPO baseline.}
  \label{tab:train_config}
\end{table*}

\subsection{Methodological Definitions}
\label{sec:app_definitions}

This section details the criteria and scoring functions (Tables \ref{tab:action_criteria}, \ref{tab:atomic_scoring}) and analyzes the training dynamics shown in Figure \ref{fig:appendix_actions}.

% 将两个表格放在这里，可以使用 [h] 或 [t] 让它们自然流动
\begin{table}[h]
    \centering
    \scriptsize 
    \setlength{\tabcolsep}{2pt} 
    \resizebox{\columnwidth}{!}{%
    \begin{tabular}{ll@{\hspace{0.1cm}}l} 
    \toprule
    \textbf{Category} & \textbf{Action} & \textbf{Validity Criteria ($\mathcal{C}_{valid}$)} \\ 
    \midrule
    \textbf{Coord.} & Click/Long & $\text{Dist}(p_{\text{pred}}, p_{\text{gt}}) < \epsilon_{\text{pos}}$ \\
    \midrule
    \textbf{Hybrid} & Scroll & $\text{Dist}(p_{s}) < \epsilon_{pos} \land \text{Dir}_{\surd}$ \\ 
    \midrule
    \multirow{2}{*}{\textbf{Text}} & Type & $\text{F1}(\text{txt}_{pred}, \text{txt}_{gt}) > \delta_{text}$ \\
     & Launch & $\text{Sim}(\text{app}_{pred}, \text{app}_{gt}) > 0.9$ \\
    \midrule
    \textbf{System} & Wait/Back & \multirow{2}{*}{$\mathbb{I}(\text{Type}_{pred} = \text{Type}_{gt})$} \\
    \textbf{\& Ctrl} & Home/etc. & \\
    \bottomrule
    \end{tabular}
    }
    \vspace{-0.2cm}
    \caption{Validity assessment criteria.}
    \label{tab:action_criteria}
\end{table}

% Table 6: Scoring Mechanism
\begin{table}[h]
    \centering
    \scriptsize
    \renewcommand{\arraystretch}{1.1} 
    \resizebox{\columnwidth}{!}{%
    \begin{tabular}{l|c|l}
    \toprule
    \textbf{Action} & \textbf{Metric} & \textbf{Scoring Function ($s_{raw}$)} \\ 
    \midrule
    \textbf{Click} & Spatial & $e^{-\frac{\|p_{pred} - p_{gt}\|^2}{2\sigma^2}}$ \\ 
    \midrule
    \textbf{Scroll} & Hybrid & $e^{-\frac{\|p_{s} - \hat{p}_{s}\|^2}{2\sigma^2}} \cdot \mathbb{I}(\text{Dir}_{\surd})$ \\ 
    \midrule
    \textbf{Type} & F1 & $2 \cdot \frac{\text{Pre} \cdot \text{Rec}}{\text{Pre} + \text{Rec}}$ \\ 
    \midrule
    \textbf{Launch} & Threshold & $\mathbb{I}(\text{Sim}(\text{App}_{pred}, \text{App}_{gt}) > 0.9)$ \\ 
    \midrule
    \textbf{System} & Exact & $\mathbb{I}(\text{Type}_{pred} = \text{Type}_{gt})$ \\ 
    \bottomrule
    \end{tabular}
    }
    \vspace{-0.2cm}
    \caption{Atomic action scoring mechanism.}
    \label{tab:atomic_scoring}
\end{table}

\subsection{Analysis of Remaining Action Primitives}
\label{sec:app_action_ablation}

Figure \ref{fig:appendix_actions} summarizes training dynamics for the remaining six action primitives.
Consistent with \textit{PressBack}, SOLAR-RL exhibits improved stability across diverse action types.
Notably, the gains are most pronounced for primitives that can trigger irreversible state changes (e.g., \textit{Launch}) or amplify downstream compounding errors (e.g., \textit{Scroll} and \textit{Finished}).
This trend aligns with SOLAR-RL's design: once a rollout enters an invalid state, failure-point based credit assignment prevents misleading rewards from propagating to earlier steps, and target-aligned shaping further stabilizes the return scale across trajectories.
In contrast, primitives with denser local supervision (e.g., \textit{Click}) naturally leave less room for improvement, leading to smaller margins.
Overall, these results suggest that SOLAR-RL primarily improves long-horizon credit assignment rather than merely smoothing short-horizon rewards.

\begin{itemize}
    \setlength{\itemsep}{1pt}
    \setlength{\parskip}{0pt}
    \item \textbf{System transitions (Launch).} In Figure \ref{fig:launch}, SOLAR-RL achieves comparable or higher reward with lower variance, indicating more reliable state-switch decisions.
    \item \textbf{Termination (Finished).} In Figure \ref{fig:finished}, the baseline collapses late in training (reward drop), while SOLAR-RL remains robust, indicating better long-horizon termination awareness.
    \item \textbf{Long-horizon control (Scroll).} Figure \ref{fig:scroll} shows reduced oscillation for SOLAR-RL, suggesting fewer redundant/overshooting scrolls that can derail downstream steps.
    \item \textbf{Temporal primitive (Wait).} Figure \ref{fig:wait} shows smoother learning for SOLAR-RL, consistent with discouraging unproductive waiting and idle loops.
    \item \textbf{Text-conditioned (Type).} In Figure \ref{fig:type}, SOLAR-RL is steadier, suggesting reduced reward noise for partially-correct typings.
    \item \textbf{Fine-grained pointing (Click).} Figure \ref{fig:click} shows similar or slightly better reward with lower variance; gains are smaller since click is more densely supervised.
\end{itemize}

\subsection{Failure Case Analysis: Continuous Decision Recovery in Long-Horizon GUI Tasks}
\label{sec:app_case_study}

\begin{figure*}[t]
  \centering
  \includegraphics[width=0.98\textwidth]{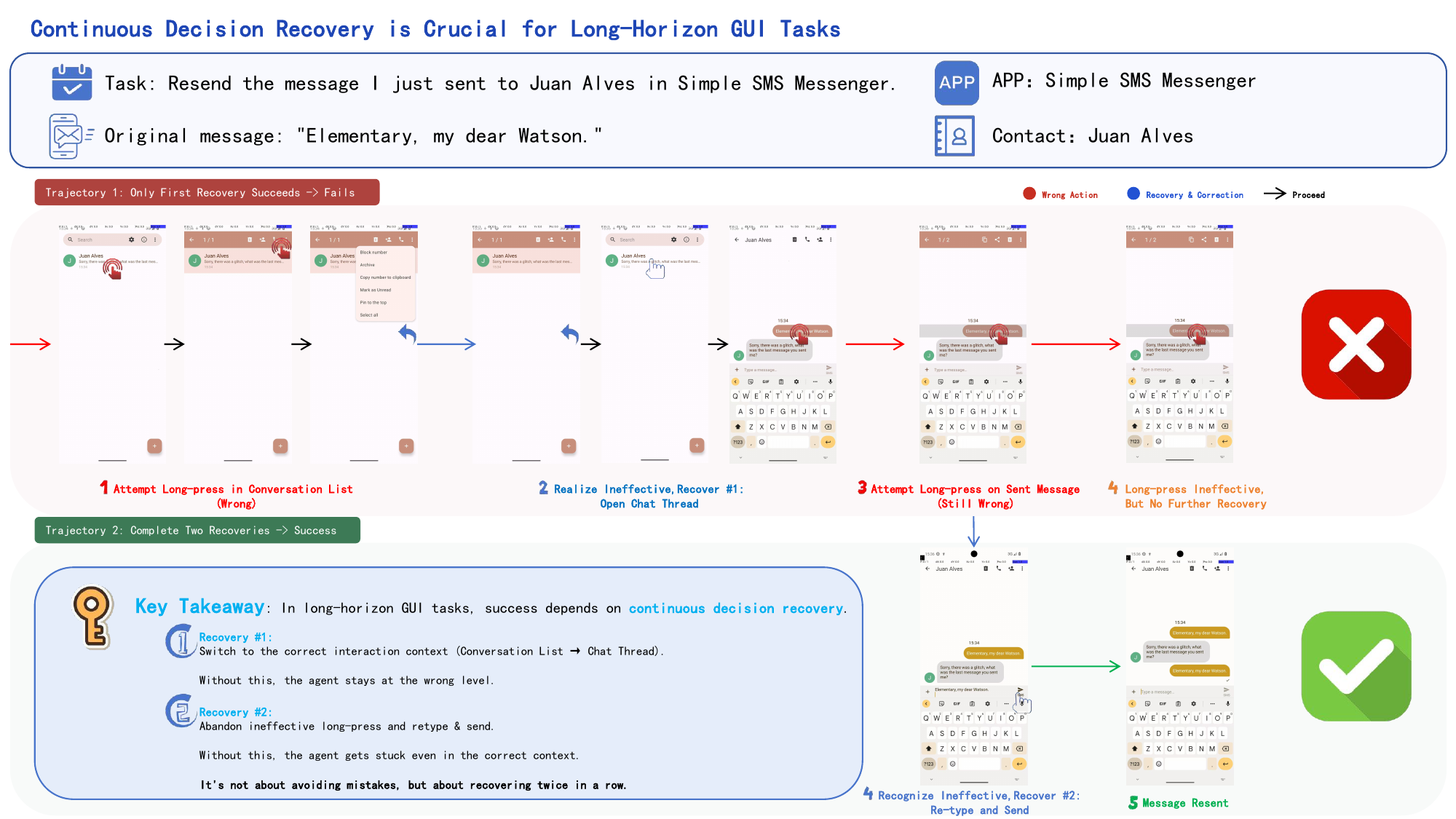}
  \caption{A qualitative failure-case study on continuous decision recovery. Both trajectories start from the same task, but
only the second one succeeds because the agent performs two consecutive recoveries: it first switches to the correct interaction
context and then replaces an ineffective long-press strategy with re-type-and-send. The example illustrates that in long-horizon
GUI tasks, robustness depends not only on avoiding errors, but on recovering from multiple errors in sequence.}
  \label{fig:case_study}
\end{figure*}

Figure~\ref{fig:case_study} presents a representative case from training on \textit{Simple SMS Messenger}. The task is to resend the most recent message to a specific contact. This example highlights that long-horizon success does not merely depend on avoiding the first mistake; rather, it depends on whether the agent can recover repeatedly after entering suboptimal states.

In Trajectory 1, the agent first makes an incorrect context-level decision by attempting a long-press in the conversation list instead of entering the chat thread. It then partially recovers by opening the correct thread, but fails to execute the second recovery: after realizing that long-pressing the sent message is ineffective, it does not switch to the correct alternative strategy of re-typing and re-sending the message. As a result, the rollout remains trapped in a locally plausible but globally unsuccessful behavior pattern.

Trajectory 2 follows a similar early prefix but successfully performs two consecutive recoveries. The agent first switches to the correct interaction context (conversation list $\rightarrow$ chat thread), and then revises the action strategy (ineffective long-press $\rightarrow$ re-type-and-send). This leads to successful task completion. The comparison shows that long-horizon GUI tasks require \emph{continuous decision recovery}: correcting one mistake is often insufficient if the agent cannot adapt again when the
next local strategy fails.

This phenomenon is consistent with the design motivation of SOLAR-RL. Failure-point based credit assignment helps localize where the rollout first becomes invalid, while trajectory-aware reward shaping discourages persistent but unproductive action chains after the breakdown point. Qualitatively, this encourages the policy to abandon ineffective local behaviors and re-enter a valid decision path instead of repeating superficially similar actions.

\subsection{Detailed Ablation on Trajectory Lengths}
\label{sec:app_length_ablation}

\begin{figure}[h]
    \centering
    \includegraphics[width=0.85\columnwidth]{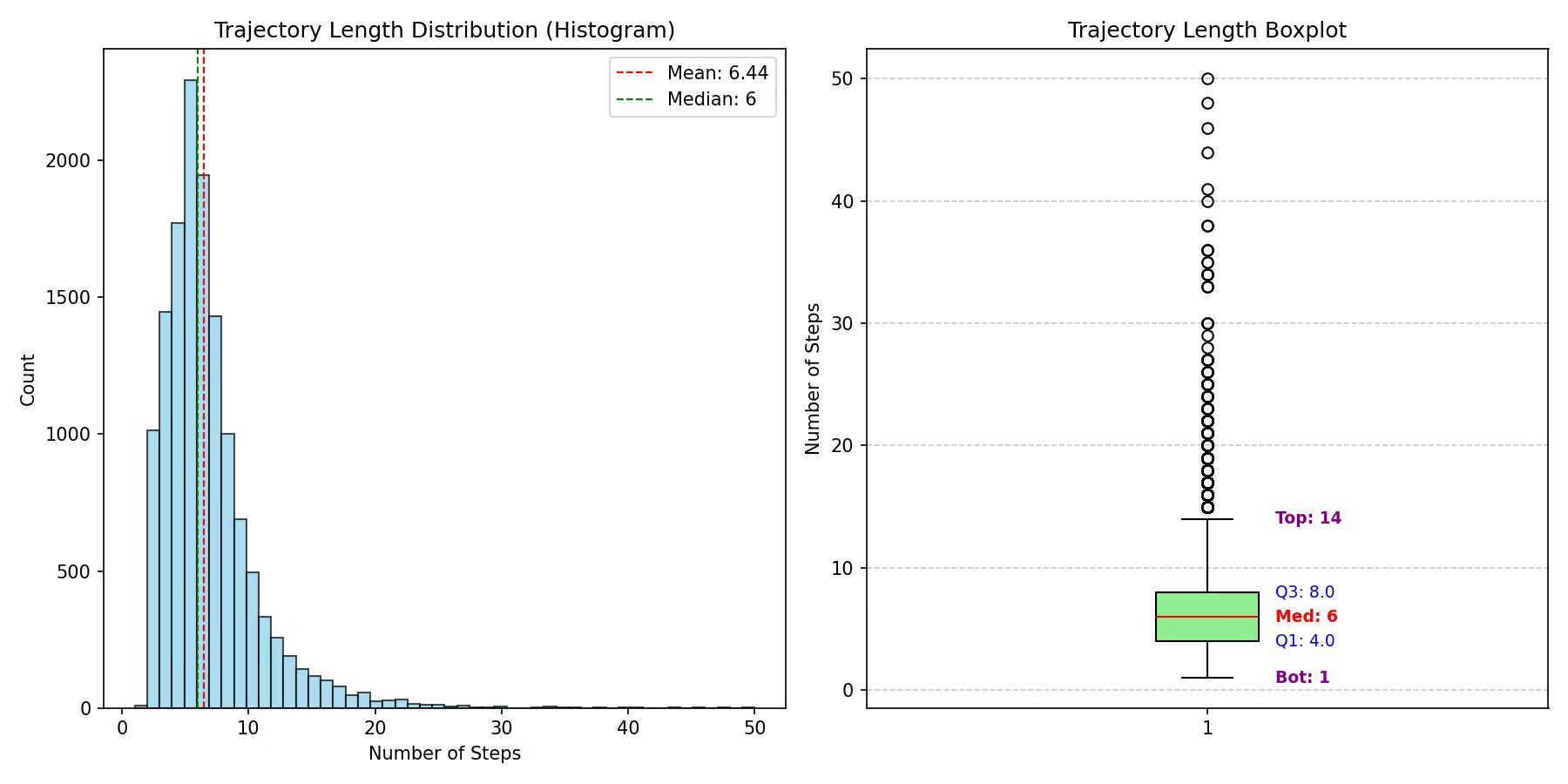}
    \caption{Distribution of trajectory lengths in the training dataset. We use the distribution statistics to define the "Long" and "Super Long" buckets for ablation.}
    \label{fig:traj_stats}
\end{figure}

In Section \ref{sec:ablation_direct}, we partitioned tasks based on the statistical distribution of optimal path lengths in the training data. Figure \ref{fig:traj_stats} presents the histogram and boxplot of these lengths. Based on the quartiles ($Q1=4, Median=6, Q3=8$), we define "Long" tasks as the interquartile extension ($L \in [6, 13]$) and "Super Long" tasks as the upper tail ($L \ge 14$).

\begin{figure}[!ht]
    \centering
    \captionsetup{font=small}
    \captionsetup[subfigure]{font=small,skip=1pt}

    % ---------- Row 1: Long ----------
    \begin{subfigure}[b]{0.49\textwidth}
        \centering
        \includegraphics[width=\linewidth,height=0.20\textheight,keepaspectratio]{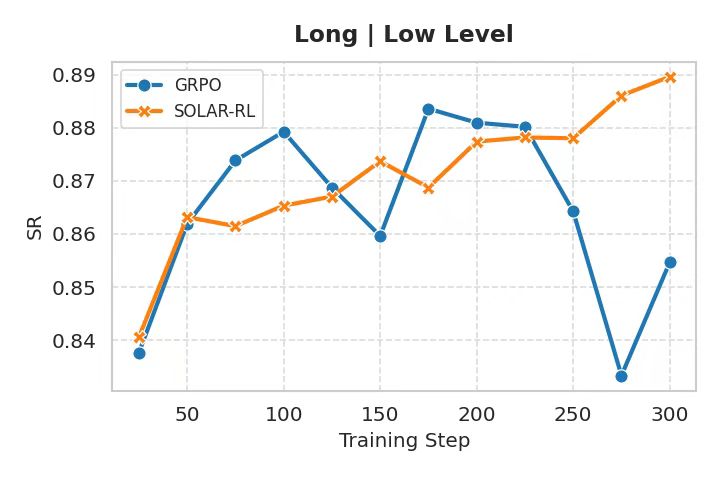}
        \caption{Low Level | Long}
        \label{fig:low_long}
    \end{subfigure}\hfill
    \begin{subfigure}[b]{0.49\textwidth}
        \centering
        \includegraphics[width=\linewidth,height=0.20\textheight,keepaspectratio]{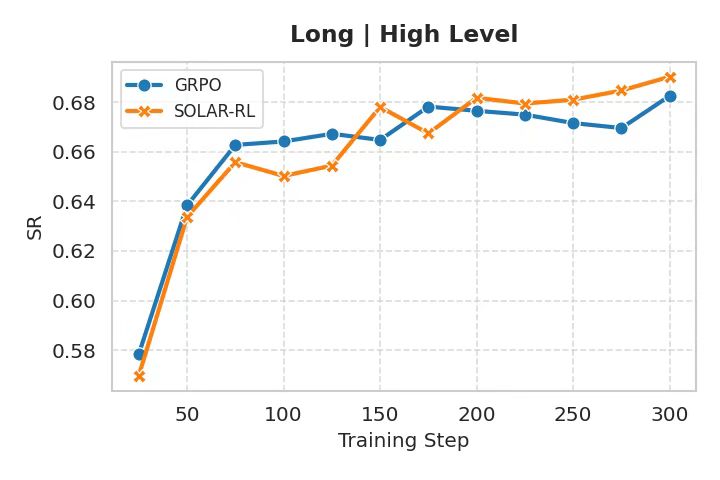}
        \caption{High Level | Long}
        \label{fig:high_long}
    \end{subfigure}

    % ---------- Row 2: Super Long ----------
    \begin{subfigure}[b]{0.49\textwidth}
        \centering
        \includegraphics[width=\linewidth,height=0.20\textheight,keepaspectratio]{latex/images/1_stage_low_Super_Long_SR.png}
        \caption{Low Level | Super Long}
        \label{fig:low_super_long}
    \end{subfigure}\hfill
    \begin{subfigure}[b]{0.49\textwidth}
        \centering
        \includegraphics[width=\linewidth,height=0.20\textheight,keepaspectratio]{latex/images/1_stage_high_Super_Long_SR.png}
        \caption{High Level | Super Long}
        \label{fig:high_super_long}
    \end{subfigure}

    \vspace{-2mm}
    \caption{\textbf{Complete direct-training ablation (Action SR).}
    Long: $L\in[6,13]$; Super Long: $L\ge 14$. SOLAR-RL consistently outperforms GRPO, with a larger gap on longer and harder horizons.}
    \label{fig:app_long_ablation}
\end{figure}

Figure \ref{fig:app_long_ablation} provides the full set of training dynamics comparisons, including the "Long" bucket ($L \in [6, 13]$) which was omitted from the main text. Consistent with the Super-Long results, SOLAR-RL (Orange) demonstrates superior sample efficiency and stability across both difficulty levels (Low/High) in this intermediate length range.

\FloatBarrier

\end{document}